\documentclass{article}

 \usepackage[nonatbib,preprint]{neurips_2026}

\usepackage[ruled]{algorithm2e} 

\usepackage{amsmath}
\usepackage{mathtools}
\usepackage{amsthm}
\usepackage{amssymb}
\usepackage{booktabs}
\usepackage{multirow}
\usepackage{graphicx}
\usepackage{url}
\usepackage{hyperref}

\usepackage{media9}
\usepackage{siunitx}
\usepackage{pifont}
\usepackage{lipsum}
\usepackage{amsmath}
\usepackage{stfloats}
\usepackage{enumitem}
\usepackage{tcolorbox}
\usepackage{ragged2e}
\usepackage{subcaption}
\usepackage[percent]{overpic}
\usepackage{xcolor}
\usepackage{wrapfig}

\AtBeginDocument{
  \setlength{\abovedisplayskip}{4pt}
  \setlength{\belowdisplayskip}{4pt}
  \setlength{\abovedisplayshortskip}{2pt}
  \setlength{\belowdisplayshortskip}{2pt}
}

\newcommand\mypara[1]{\vspace{1pt}\noindent\textbf{#1.}}

\title{Escaping Confidence Trap: Evolutionary Decoding for Mathematical Reasoning in Diffusion LLMs}

\author{%
  Zhenhong Sun\thanks{Work done during the visit to Nanyang Technological University and National University of Singapore.} \\
  Australian National University \\
  Canberra, Australia \\
  \texttt{zhenhong.sun@anu.edu.au}
  \And
  Hanqing Zhao \\
  Nanyang Technological University \\
  Singapore \\
  \texttt{zhaohanqing.ac@gmail.com}
  \AND
  Yatao Bian \\
  National University of Singapore \\
  Singapore \\
  \texttt{ybian@nus.edu.sg}
  \And
  Rong-Cheng Tu \\
  Nanyang Technological University \\
  Singapore \\
  \texttt{turongcheng@gmail.com}
  \AND
  Liuyue Xie \\
  Amazon \\
  Seattle, WA, USA \\
  \texttt{liuyuex@andrew.cmu.edu}
  \And
  Xu Zhang \\
  Amazon\\
  Seattle, WA, USA \\
  \texttt{xzhnamz@amazon.com}
  \AND
  Jue Wang \\
  Amazon \\
  Seattle, WA, USA \\
  \texttt{juewangn@amazon.com}
  \And
  Davide Modolo \\
  Amazon \\
  Seattle, WA, USA \\
  \texttt{dmodolo@amazon.com}
  \AND
  Daoyi Dong \\
  University of Technology Sydney \\
  Sydney, Australia \\
  \texttt{daoyidong@gmail.com}
  \And
  Dacheng Tao \\
  Nanyang Technological University \\
  Singapore \\
  \texttt{dacheng.tao@ntu.edu.sg}
}

\begin{document}

\maketitle

\begin{center}
  \small
  \textbf{Project Page:}
  \href{https://engineeringai-lab.github.io/evolutionary-decoding}
       {\texttt{https://engineeringai-lab.github.io/evolutionary-decoding}}
    \vspace{15pt}
\end{center}

\begin{abstract}
Diffusion large language models (dLLMs) have emerged as a promising alternative to autoregressive LLMs, offering efficient generation through block-wise progressive unmasking. 
However, their strong general-purpose performance does not necessarily translate into reliable mathematical reasoning, where correctness depends on preserving coherent numerical-symbolic reasoning trajectories. 
In this work, we analyze the decoding trajectories of LLaDA 2.0 and identify a recurring diffusion confidence trap: local token confidence can become misaligned with global reasoning correctness during progressive block decoding. 
Our analysis reveals two representative failure regimes: sampling-sensitive failures, where correct paths exist but are unstable, and sampling-consistent failures, where repeated sampling converges to repetitive high-confidence but incorrect continuations. 
Motivated by this observation, we propose Evolutionary Decoding, a training-free test-time scaling framework that views diffusion decoding as an evolutionary process over candidate reasoning states. 
The framework combines step-wise selection, which preserves useful numerical-symbolic signals and suppresses repetitive patterns, with block-wise mutation, which introduces structured alternatives to escape incorrect high-confidence basins. 
Experiments on multiple benchmarks show that Evolutionary Decoding improves LLaDA 2.0 over confidence-based decoding, leading to more reliable mathematical reasoning.
\end{abstract}

\section{Introduction}

Diffusion large language models (dLLMs) are emerging as a strong challenger to the dominant autoregressive generation paradigm~\cite{nie2025large,bie2025llada2,ye2025dream}. 
By generating text through iterative denoising and progressively determining masked tokens, dLLMs provide efficient and flexible generation while showing competitive general-purpose language modeling ability. 
However, strong general performance does not necessarily translate into reliable behavior in precision-demanding domains, such as mathematical reasoning, legal analysis, and scientific problem solving, where correctness depends on preserving critical intermediate structures rather than producing locally plausible continuations~\cite{ye2024diffusion,ye2024beyond,chen2025reasoning}. 
Mathematical reasoning provides a representative testbed for this challenge, because solving a problem often requires a continuous block of numbers, symbols, and transformations to be generated consistently, rather than producing isolated fluent tokens. 
This naturally raises our central question: \textit{can the block-wise progressive decoding of dLLMs better support mathematical reasoning by preserving coherent numerical-symbolic reasoning blocks?}

Recent efforts have begun to improve the mathematical reasoning ability of diffusion LLMs from several directions. 
One line of work is to enhance the base model through large-scale instruction tuning or domain-specific supervised fine-tuning, enabling dLLMs to better imitate mathematical solution formats and chain-of-thought style reasoning~\cite{ye2024diffusion,zhu2025llada}. 
Another line is to incorporate preference optimization, reinforcement learning, or verifier-based feedback to align generated solutions with final-answer correctness or intermediate reasoning quality~\cite{cobbe2021training,zhao2025d1,zhu2025dirl}. 
A third direction focuses on inference-time strategies, such as repeated sampling~\cite{wang2022self,kang2025scalablebestN}, confidence-based decoding~\cite{cai2026confidence,fang2026locally}, remasking~\cite{wang2503remasking}, reward-free guidance~\cite{chen2025rfg}, or test-time search~\cite{shen2026improving,bai2026prism,bilal2026s,lee2025testmath}, aiming to exploit the stochasticity and parallel prediction structure of diffusion generation. 
While these methods improve the overall reasoning performance of dLLMs, they also motivate a closer examination of the internal trajectory patterns induced by confidence-driven progressive unmasking, especially how numerical and symbolic tokens are formed, preserved, or lost during block-wise decoding.

In this paper, we analyze the decoding trajectories of LLaDA 2.0 on mathematical reasoning tasks. 
If dLLM failures were mainly caused by sampling noise, repeated sampling would be expected to recover correct reasoning paths. 
However, eight independent runs per problem reveal that many errors persist across samples, suggesting that the limitation is more systematic and is tied to confidence-driven block-wise decoding.
We observe two representative regimes. 
In \textbf{sampling-sensitive failures}, some trajectories succeed while others fail, indicating that correct paths exist but are not reliably preserved during decoding. 
These failures often occur when the model transitions from natural-language scaffolding to numerical-symbolic reasoning, where critical digits or operators may be delayed or suppressed. 
In \textbf{sampling-consistent failures}, all trajectories fail and often converge to similar incorrect continuations, showing that the decoder repeatedly enters a high-confidence but low-value reasoning basin. 
Together, these results reveal a \textbf{diffusion confidence trap}: local token confidence can become misaligned with global mathematical correctness, either by failing to preserve useful uncertain tokens or by confidently committing to repetitive low-value blocks.

These findings motivate us to view confidence-guided diffusion decoding as an evolutionary process over candidate reasoning states, where local confidence drives token growth within blocks, completed blocks inherit and evolve the reasoning trajectory, and the two failure regimes arise from unstable selection of numerical-symbolic signals and insufficient mutation away from repetitive attractors.
Motivated by this analysis, we propose \textbf{Evolutionary Decoding}, a training-free test-time scaling framework for mathematical reasoning in diffusion LLMs. 
Instead of modifying model parameters or relying on more independent samples, our method intervenes in the internal evolution of block-wise diffusion decoding. 
Specifically, \textbf{step-wise selection} adjusts which tokens survive during progressive unmasking by enhancing moderately uncertain numerical-symbolic tokens and suppressing repetitive block patterns, thereby improving the preservation of reasoning-critical signals. 
Meanwhile, \textbf{block-wise mutation} perturbs low-support decoding paths with structured numerical and symbolic alternatives, expanding the candidate reasoning population before the block collapses into a fixed direction. 
In essence, our method changes how reasoning states are selected, preserved, and diversified during test-time decoding, allowing dLLMs to mitigate sampling-sensitive failures and escape sampling-consistent high-confidence traps. 
Experiments on multiple datasets show that Evolutionary Decoding improves LLaDA 2.0 over confidence-based decoding, demonstrating more reliable mathematical reasoning and stronger test-time scaling behavior.

The main contributions of this paper are summarized as follows:
\begin{itemize}[leftmargin=*, noitemsep, nolistsep]
    \item[$\bullet$] We analyze mathematical sampling trajectories in dLLMs and interpret the \textbf{diffusion confidence trap} into two failure regimes from an evolutionary view of block-wise reasoning states.

    \item[$\bullet$] We propose \textbf{step-wise selection} to address sampling-sensitive failures by preserving uncertain but useful numerical-symbolic tokens during progressive unmasking, improving reasoning signals.

    \item[$\bullet$] We introduce \textbf{block-wise mutation} to address sampling-consistent failures by diversifying low-support reasoning paths before they collapse into repetitive high-confidence traps.
\end{itemize}

\section{Related Work}

\mypara{dLLMs and Mathematical Reasoning}
Diffusion language models generate text through iterative denoising or masked-token refinement, rather than left-to-right autoregression. 
Early work established the foundations of discrete and masked diffusion language modeling, including structured denoising in discrete state spaces~\cite{austin2021structured}, likelihood-based diffusion language modeling~\cite{gulrajani2023likelihood}, ratio-estimation-based discrete diffusion~\cite{lou2023discrete}, simplified masked diffusion for discrete data~\cite{shi2024simplified}, and simple masked diffusion language modeling~\cite{sahoo2024simple}. 
Subsequent studies scaled masked diffusion to text~\cite{nie2024scaling} and explored block diffusion as a bridge between parallel refinement and semi-autoregressive generation~\cite{arriola2025block}. 
More recently, large-scale diffusion LLMs such as LLaDA~\cite{nie2025large}, LLaDA1.5~\cite{zhu2025llada}, LLaDA2.0~\cite{bie2025llada2}, Mercury~\cite{khanna2025mercury}, Dream~\cite{ye2025dream}, and sparse MoE diffusion variants~\cite{zhu2025lladamoe} have shown strong general-purpose generation ability. 
For math-related reasoning, Diffusion-of-Thought supports chain-of-thought reasoning through iterative refinement~\cite{ye2024diffusion}, while discrete diffusion has also been explored for complex reasoning and planning~\cite{ye2024beyond}, and reasoning may concentrate in dynamic confusion zones~\cite{chen2025reasoning}, producing locally plausible but globally incorrect trajectories. In this work, we select LLaDA2.0 as a representative baseline for studying mathematical reasoning.

\mypara{Fine-tuning and Test-time Scaling for Reasoning}
Fine-tuning and post-training methods improve mathematical reasoning by updating the model or its preference behavior, such as verifier training~\cite{cobbe2021training}, reinforcement learning for diffusion LLM reasoning~\cite{zhao2025d1}, efficient diffusion post-training~\cite{zhu2025dirl}, and variance-reduced preference optimization for diffusion LLMs~\cite{zhu2025llada}. 
In parallel, test-time scaling methods improve reasoning without modifying model parameters, including chain-of-thought prompting~\cite{wei2022chain}, self-consistency~\cite{wang2022self}, optimal allocation of test-time computation~\cite{snell2024scaling}, and scalable best-of-\(N\) selection via self-certainty~\cite{kang2025scalablebestN}. 
Recent work further extends test-time scaling to diffusion LLMs through remasking~\cite{wang2503remasking}, reward-free guidance~\cite{chen2025rfg}, joint search over generation order and token space~\cite{shen2026improving}, hidden semi-autoregressive experts~\cite{lee2025testmath}, hierarchical search with self-verification~\cite{bai2026prism}, and stratified scaling search~\cite{bilal2026s}. 
Other studies analyze confidence-based decoding from complementary perspectives, showing its theoretical efficiency~\cite{cai2026confidence}, its quality-exploration dilemma~\cite{fang2026locally}, and block-diffusion test-time scaling behavior~\cite{lu2026advancingbacdtccf}. 
Our method performs structured training-free mutation and selection within a single diffusion trajectory.

\mypara{Evolutionary Algorithms}
Evolutionary algorithms provide a general optimization paradigm based on population, variation, selection, and survival, and have been widely used for search problems where direct gradient-based optimization is difficult~\cite{eiben2003introduction,de2017evolutionary}. Recent work has connected evolutionary computation with LLMs, mainly using LLMs to generate, mutate, or evaluate prompts and candidate solutions~\cite{guo2023connecting,chao2024large}. 
Unlike prior works that use LLMs as external evolutionary operators, we view diffusion decoding itself as an evolutionary process, where block-level reasoning states compete, mutate toward numerical-symbolic alternatives, and are selected to escape confidence traps.


\section{Methodology}

\subsection{Challenge of Confidence Trap}
\label{sec:confidence_trap}

\mypara{dLLM Preliminary}
Diffusion LLMs such as LLaDA generate responses through a block-wise autoregressive process. 
Given a prompt $\mathbf{x}$, the response is divided into consecutive blocks 
$\mathbf{y}=(\mathcal{B}_1,\ldots,\mathcal{B}_M)$, which are generated sequentially from left to right. 
When decoding the $m$-th block, all previously completed blocks $\mathbf{y}_{<m}$ are fixed as context, while the current block $\mathcal{B}_m$ is initialized with masked tokens. 
The model then performs a step-wise diffusion process inside this block: at each step, it re-predicts all still-masked positions in parallel according to the current partially revealed block state.
Formally, at diffusion step $t$, for each masked position $i\in\mathcal{B}_m$, LLaDA predicts a candidate token and assigns it a confidence score:
\begin{equation}
\hat{y}_i^{(t)}=\arg\max_{v\in\mathcal{V}} 
p_\theta(v \mid \mathbf{x}, \mathbf{y}_{<m}, \mathcal{B}_m^{(t)}),
\quad
c_i^{(t)}=\max_{v\in\mathcal{V}} 
p_\theta(v \mid \mathbf{x}, \mathbf{y}_{<m}, \mathcal{B}_m^{(t)}),
\end{equation}
where $\mathcal{V}$ denotes the vocabulary and $v\in\mathcal{V}$ is a candidate token, and positions with $c_i^{(t)} \geq \tau$ are unmasked and fixed into the block, while the remaining positions stay masked and are reconsidered in the next diffusion step. 
This process repeats until the current block is completed; the completed block is then appended for next block decoding. 
Therefore, LLaDA decoding can be viewed as autoregressive block prediction, coupled with confidence-driven diffusion unmasking per block.


\begin{figure}[t]
    \centering
        \begin{minipage}[t]{0.48\textwidth}
        \vspace{0pt}
        \centering

        \begin{overpic}[width=\linewidth]{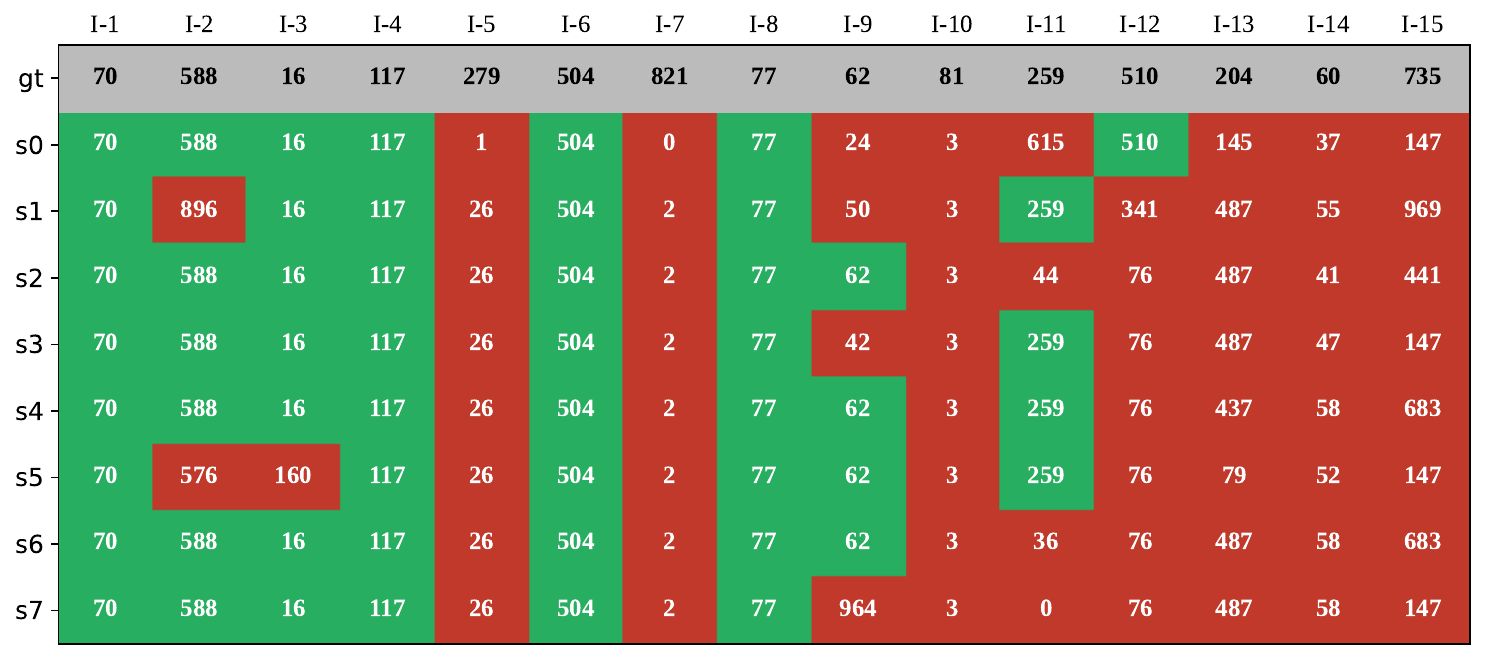}
        \put(0,42){{\scriptsize\bfseries (a)}}
        \end{overpic}
        \vspace{-10pt}

        \begin{overpic}[width=\linewidth]{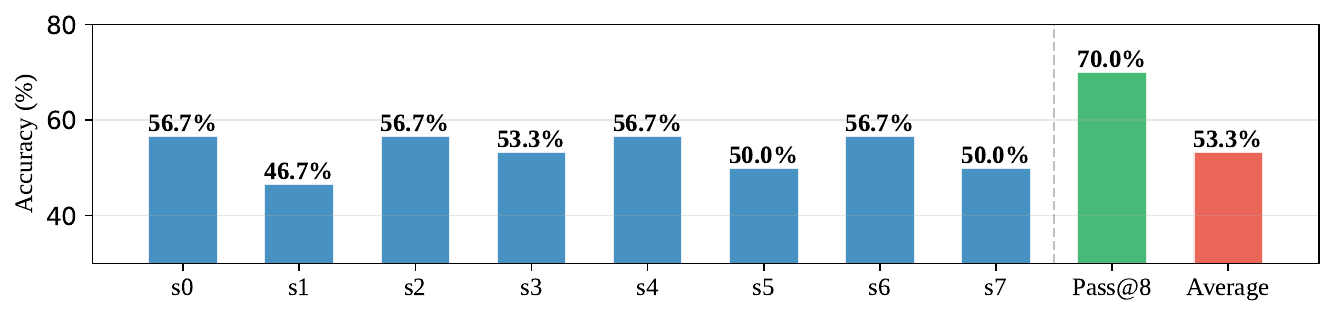}
        \put(0,19){{\scriptsize\bfseries (b)}}
        \end{overpic}
        \vspace{-10pt}
        
        \begin{overpic}[width=\linewidth]{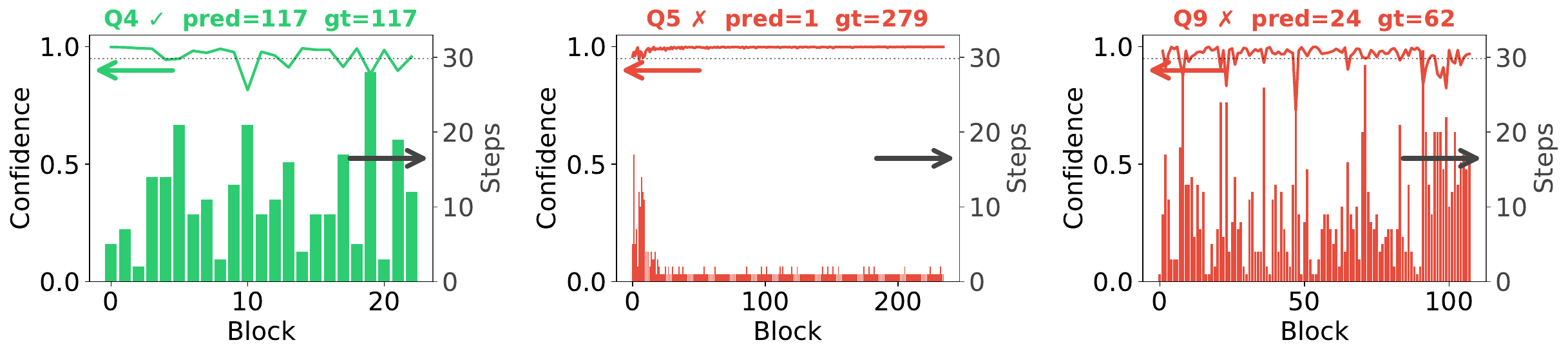}
        \put(0,21){{\scriptsize\bfseries (c)}}
        \put(40,15){{\fontsize{5}{6}\selectfont\bfseries Consistent Failure}}
        \put(75,15){{\fontsize{5}{6}\selectfont\bfseries Sensitive Failure}}
        \end{overpic}

    \end{minipage}
    \hfill
    \begin{minipage}[t]{0.49\textwidth}
        \vspace{0pt}
        \centering

        \begin{overpic}[width=\linewidth]{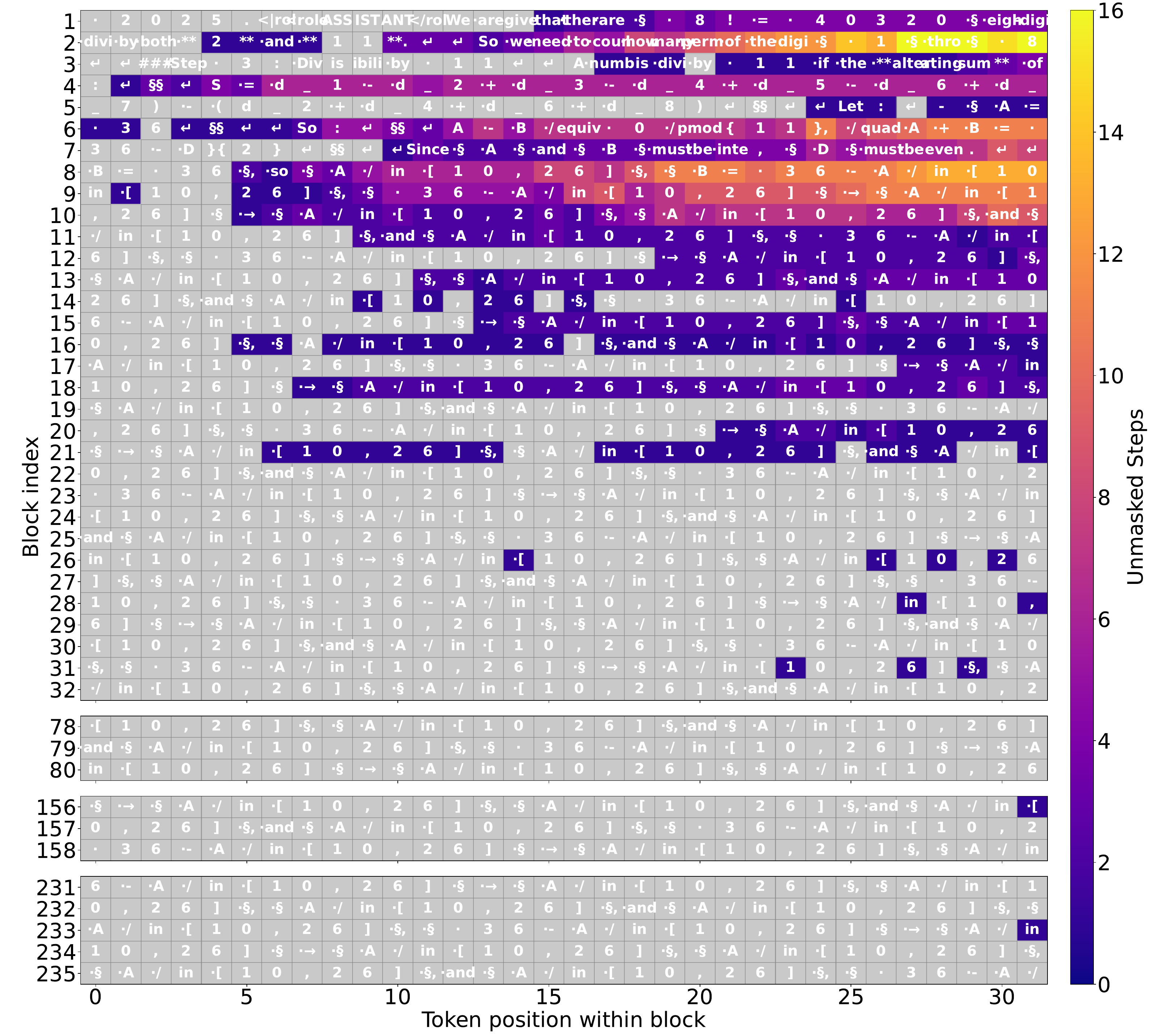}
        \put(0,87){{\scriptsize\bfseries (d)}}
        \put(8,25){{\fontsize{5}{6}\selectfont\bfseries Q5 Consistent Failure: high-confidence mass unmasking by repetitive tokens.}}
        \end{overpic}

    \end{minipage}
    \caption{
    The \textbf{diffusion confidence trap} in LLaDA 2.0. 
    (a) Eight independent runs reveal two failure regimes: sampling-consistent failures and sampling-sensitive failures. 
    (b) Pass@8 remains bounded at 70.0\%, indicating that repeated sampling alone cannot solve all failures. 
    (c) Block-wise statistics show distinct trajectory patterns for successful, sampling-consistent, and sampling-sensitive cases. 
    (d) A representative sampling-consistent failure further illustrates the transition from step-wise unmasking to high-confidence repetition. Gray boxes are 1-step unmasked tokens.
    }
    \label{fig:confidence_trap_analysis}
\end{figure}
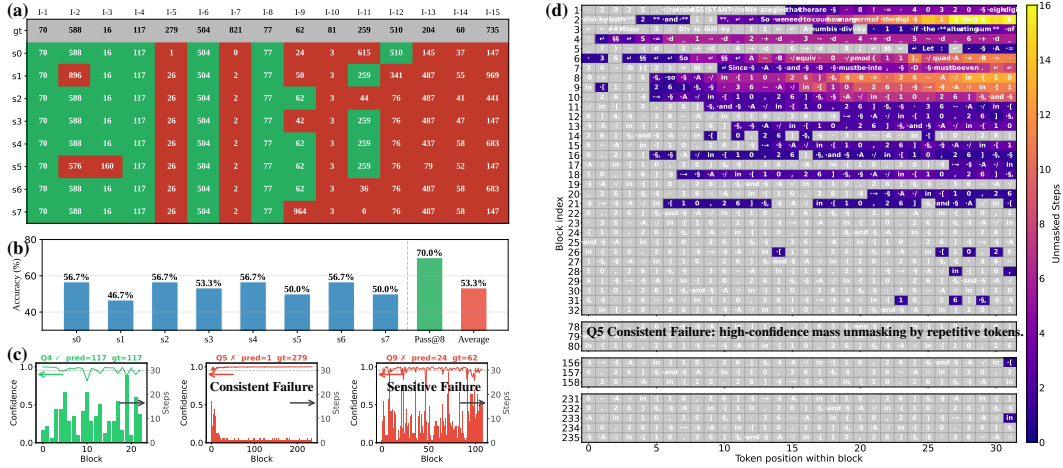

\mypara{Diffusion Confidence Trap}
Based on the above block-wise diffusion mechanism, we next examine whether repeated sampling can resolve LLaDA's failures. 
If errors mainly arise from stochastic variation, multiple independent runs should recover correct trajectories with high probability. 
We therefore run LLaDA 2.0 with SGLang~\cite{zheng2024sglang} eight times per problem and analyze both final correctness and intermediate decoding trajectories in Figure~\ref{fig:confidence_trap_analysis}. 
Figure~\ref{fig:confidence_trap_analysis}(a) reveals two regimes: \textbf{sampling-consistent failures}, where all trajectories fail, and \textbf{sampling-sensitive failures}, where only some trajectories succeed. 
However, pass@8 reaches only 70.0\% in Figure~\ref{fig:confidence_trap_analysis}(b), leaving around 30.0\% of problems unsolved across all attempts. 
This indicates that repeated sampling is insufficient and points to a systematic limitation in confidence-driven decoding.

We further inspect the failed trajectories to understand how the decoding process deviates. 
As shown in Figure~\ref{fig:confidence_trap_analysis}(c), successful trajectories are confident and efficient, completing reasoning with fewer blocks and fewer diffusion steps while maintaining compact symbolic or numerical derivations. 
By contrast, \textbf{sampling-consistent failures} contain more redundant natural-language filler together with high digit/symbol confidence, suggesting repeated convergence to superficially confident but low-value reasoning paths. 
Figure~\ref{fig:confidence_trap_analysis}(d) further illustrates this process: decoding initially follows a step-wise autoregressive pattern, revealing connected natural-language tokens first and gradually exposing symbolic or numerical reasoning tokens, but eventually falls into a high-confidence mass-unmasking stage dominated by repetitive continuations. 
Meanwhile, \textbf{sampling-sensitive failures} exhibit stronger fluctuations in inference steps and lower symbol/digit density, indicating that the model may deviate during uncertain transitions between natural-language scaffolding and symbolic reasoning. 
In these cases, correct reasoning paths may exist, but the diffusion trajectory is sensitive to intermediate confidence variations and can drift into low-information continuations.
We refer to these behaviors as the \textbf{diffusion confidence trap}: local token-level confidence becomes misaligned with global reasoning correctness, either by confidently committing to repetitive low-value tokens or by allowing uncertain transitions to steer the trajectory away from useful reasoning.

\mypara{Evolutionary Decoding}
These findings motivate us to reinterpret confidence-guided diffusion decoding as an evolutionary process over a population of candidate reasoning states. 
During block-wise autoregressive generation, the LLM defines an implicit environment, while local token confidence acts as a fitness-like signal that determines which tokens are released and propagated. 
Within each block, diffusion steps correspond to the gradual maturation of candidate states; across blocks, the reasoning trajectory is inherited and evolved toward a final answer. 
A correct answer emerges when a trajectory successfully preserves useful reasoning signals, avoids repetitive attractors, and survives through successive blocks.
This perspective naturally explains the two failure regimes. 
Sampling-sensitive failures arise when the selection signal is unstable: useful numerical or symbolic tokens may be delayed or suppressed during uncertain diffusion steps, causing the trajectory to drift. 
Sampling-consistent failures arise when the population lacks effective mutation: repeated decoding follows the same high-confidence but incorrect basin, leading to repetitive low-value continuations. 

\begin{figure}[t]
    \centering
    \includegraphics[width=1\linewidth]{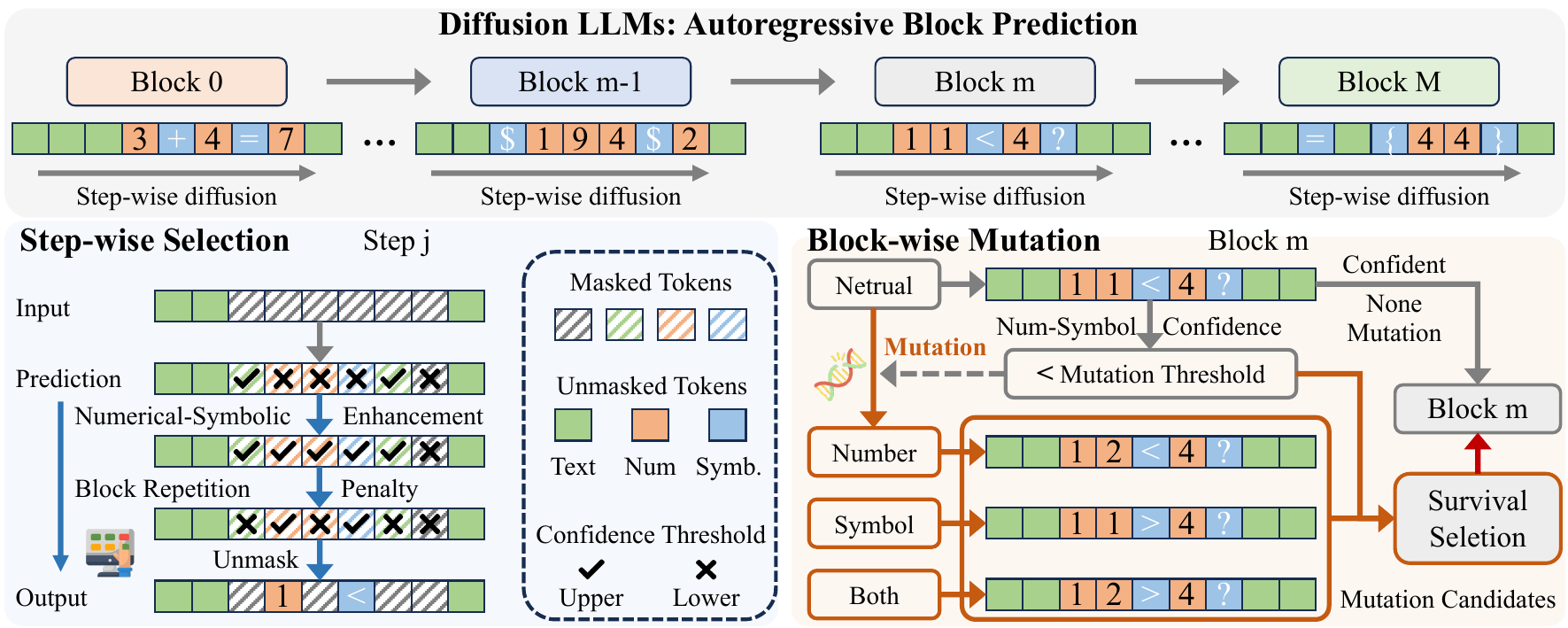}
    \caption{
    Overview of \textbf{Evolutionary Decoding} framework for mathematical reasoning in diffusion LLMs. 
    Given dLLM autoregressive block prediction, we intervene in the block diffusion process through \textbf{step-wise selection} and \textbf{block-wise mutation} to help escape the confidence trap. 
    }
    \label{fig:pipeline}
\end{figure}

Therefore, we propose an \textbf{Evolutionary Decoding} framework to address the two failure regimes through two complementary operations, as summarized in Figure~\ref{fig:pipeline} and detailed in the following subsections. 
\textbf{Step-wise selection} guides the maturation of candidate reasoning states within each block by preserving numerical-symbolic tokens that are critical for reasoning during diffusion inference. 
\textbf{Block-wise mutation} perturbs collapsed high-confidence trajectories and injects alternative reasoning directions, allowing stronger block candidates to escape repetitive attractors.

\subsection{Step-wise Selection for Sampling-sensitive Failures}
\label{subsec:selection}

From the evolutionary view, sampling-sensitive failures arise when promising reasoning trajectories exist but fail to survive unstable diffusion steps. 
Since the original decoder selects tokens mainly by local confidence, it may favor fluent natural-language tokens over less confident but reasoning-critical digits and symbols. 
These numerical-symbolic tokens can therefore be delayed or suppressed, causing the trajectory to drift into low-information continuations. 
We address this issue with step-wise selection, which preserves informative tokens and penalizes repetitive block patterns.

For the \(m\)-th block at decoding step \(t\), let \(\mathcal{M}_{m}^{t}\) denote the set of masked positions, and let \(c_{m,i}^{t}\) be the top-1 confidence of position \(i\in\mathcal{M}_{m}^{t}\). 
We define the selection score as
\begin{equation}
s_{m,i}^{t}
=
c_{m,i}^{t}
+
s_{\mathrm{en}}(m,i,t)
-
s_{\mathrm{re}}(m,t),
\qquad
x_{m,i}^{t+1}
=
\begin{cases}
\hat{x}_{m,i}^{t}, & s_{m,i}^{t}\geq \tau,\\
\texttt{[MASK]}, & s_{m,i}^{t}<\tau,
\end{cases}
\label{eq:step_selection}
\end{equation}
where \(\hat{x}_{m,i}^{t}\) is the current top-1 prediction, \(\tau\) is the release threshold, \(s_{\mathrm{en}}\) denotes the numerical-symbolic enhancement term, and \(s_{\mathrm{re}}\) denotes the block-level repetition penalty. 
Compared with the original confidence-only rule, Eq.~\eqref{eq:step_selection} keeps confidence as the main release signal, but adjusts it according to whether a token is reasoning-informative with the current block that is non-repetitive.

\mypara{Numerical-Symbolic Enhancement}
The enhancement term \(s_{\mathrm{en}}(m,i,t)\) promotes reasoning-critical numerical and symbolic tokens during step-wise unmasking. 
We instantiate it as
\begin{equation}
H_{m,i}^{t}
=
-\sum_{v}p_{m,i}^{t}(v)\log p_{m,i}^{t}(v),
\qquad
s_{\mathrm{en}}(m,i,t)
=
\alpha \eta_{m,i}^{t} H_{m,i}^{t},
\label{eq:numerical_symbolic_enhancement}
\end{equation}
where \(p_{m,i}^{t}\) is the predicted token distribution, \(\eta_{m,i}^{t}\in\{0,1\}\) indicates whether the current top-1 prediction \(\hat{x}_{m,i}^{t}\) belongs to a numerical or symbolic token set, and \(\alpha\) controls the enhancement strength. 
For ordinary natural-language tokens, \(\eta_{m,i}^{t}=0\), so the score reduces to confidence-based selection. 
For numerical or symbolic predictions, \(\eta_{m,i}^{t}=1\), allowing moderate uncertainty to increase the release score. 
This helps prevent useful but temporarily uncertain reasoning tokens from being delayed or suppressed by fluent but less informative text tokens.

\mypara{Block Repetition Penalty}
The repetition term \(s_{\mathrm{re}}(m,t)\) prevents the decoder from aggressively releasing tokens when the current block largely repeats the previous block. 
We define it as
\begin{equation}
r_{m}^{t}
=
\frac{1}{|\mathcal{C}_{m}|}
\sum_{i\in\mathcal{C}_{m}}
\mathbf{1}\!\left(
\hat{x}_{m,i}^{t}
=
\hat{x}_{m-1,i}
\right),
\qquad
s_{\mathrm{re}}(m,t)
=
\beta r_{m}^{t},
\label{eq:repetition_penalty}
\end{equation}
where \(\mathcal{C}_{m}\) is the set of comparable positions between the current and previous blocks, \(\hat{x}_{m-1,i}\) is the corresponding top-1 prediction from the previous block, and \(\beta\) controls the strength of repetition suppression. 
A larger \(r_{m}^{t}\) indicates that the current block is following the previous block too closely, which is often associated with repetitive and low-information continuation. 
Since \(s_{\mathrm{re}}(m,t)\) is shared by all masked positions in the current block, it does not change their relative ranking, but uniformly lowers their release scores when the block becomes repetitive.
A strong repetition penalty may suppress all positions even when the original decoder would have released some of them. 
We therefore use a penalty-triggered fallback. 
Let $\mathcal{M}$ denote the set of currently masked positions. 
If
\begin{equation}
\max_{i\in\mathcal{M}} c_i \ge \theta
\quad\text{but}\quad
\max_{i\in\mathcal{M}} (c_i+\alpha H_i-\beta r) < \theta,
\label{eq:penalty_trigger}
\end{equation}
we treat this as over-suppression by the selection score and reveal the top-$K$ positions ranked by the original confidence $c_i$. 
Otherwise, if the original confidence itself is below the threshold, we reveal only the most confident masked position to maintain monotonic progress. 
This fallback keeps the repetition penalty from causing persistent stalling while suppressing repetitive blocks.

\subsection{Block-wise Mutation for Sampling-consistent Failures}
\label{subsec:mutation}

While step-wise selection (Section~\ref{subsec:selection}) mitigates sampling-sensitive failures, it cannot fully address sampling-consistent failures, where trajectories repeatedly collapse to the same wrong solution. 
From an evolutionary perspective, this reflects a premature loss of diversity: once a block enters a high-confidence but flawed state, subsequent diffusion steps tend to reinforce rather than escape it. 
We therefore introduce \textbf{block-wise mutation} to inject structured numerical-symbolic alternatives during early block formation, followed by a survival rule that retains the most promising candidate.

\mypara{Structured Block-wise Mutation}
For the \(m\)-th block at decoding step \(t\), let \(\mathbf{z}_{m,i}^{t}\in\mathbb{R}^{V}\) denote the logits predicted for a masked position \(i\in\mathcal{M}_{m}^{t}\), where \(V\) is the vocabulary size. 
We globally bias the logit distribution across the entire vocabulary toward numerical and symbolic tokens, as these often determine the correctness of mathematical reasoning.
For each mutation branch \(q\), we construct the mutated logits as:
\begin{equation}
\tilde{\mathbf{z}}_{m,i}^{t,(q)}
=
\mathbf{z}_{m,i}^{t}
+
\delta
\left(
\eta_{\mathrm{num}}^{(q)}\mathbf{VM}_{\mathrm{num}}
+
\eta_{\mathrm{sym}}^{(q)}\mathbf{VM}_{\mathrm{sym}}
\right),
\qquad
i\in\mathcal{M}_{m}^{t},
\label{eq:block_mutation}
\end{equation}
where \(\mathbf{VM}_{\mathrm{num}}\) and \(\mathbf{VM}_{\mathrm{sym}}\in \{0,1\}^{V}\) are global vocabulary-level masks that identify all numerical and symbolic tokens, respectively. 
The scalar \(\delta\) controls the mutation strength, while \((\eta_{\mathrm{num}}^{(q)},\eta_{\mathrm{sym}}^{(q)})\) specifies the mutation direction. The pairs \((0,0),\ (1,0),\ (0,1),\ \text{and}\ (1,1)\) correspond to neutral, numerical, symbolic, and mixed mutation branches. 
To determine whether mutation is necessary, we first execute the neutral branch (\(\eta_{\mathrm{num}}^{(q)}=\eta_{\mathrm{sym}}^{(q)}=0\)) and compute its numerical-symbolic confidence. 
If this confidence is greater than or equal to a mutation threshold \(\tau_{\mathrm{mut}}\), the neutral trajectory is deemed sufficiently supported, and we keep only the neutral candidate. 
Otherwise, if the confidence falls below \(\tau_{\mathrm{mut}}\) and the number of mutated blocks in the current trajectory is strictly less than a global budget \(m_{\max}\), the additional mutation branches are activated. 
Together with the neutral branch, these activated branches form the candidate set \(\mathcal{Q}_m\).
Each branch in \(\mathcal{Q}_m\) then decodes the block using a shared release threshold \(\tau\). 
Specifically, the confidence of position \(i\) under branch \(q\) is:
\begin{equation}
c_{m,i}^{t,(q)}
=
\max_{v\in[V]}
\mathrm{softmax}
\left(
\tilde{\mathbf{z}}_{m,i}^{t,(q)}
\right)_v,
\qquad
\tilde{x}_{m,i}^{t+1,(q)}
=
\arg\max_{v\in[V]}
\mathrm{softmax}
\left(
\tilde{\mathbf{z}}_{m,i}^{t,(q)}
\right)_v
\quad
\text{if }
c_{m,i}^{t,(q)}\ge \tau .
\label{eq:confidence_release}
\end{equation}
This produces a candidate block state \(\tilde{\mathbf{x}}_{m}^{t+1,(q)}\) for each branch \(q\in\mathcal{Q}_m\).

\mypara{Survival Selection}
Mutation expands the local search space by generating multiple candidate block states. However, we avoid introducing an overly complex survival objective. 
During diffusion decoding, step-wise selection has already guided numerical-symbolic tokens and suppressed repetitive continuations. 
Therefore, once the candidate set \(\mathcal{Q}_m\) is generated, we simply select the candidate that maximizes the original model confidence:
\begin{equation}
q^{*}
=
\arg\max_{q\in\mathcal{Q}_m}
\bar{c}_{m}^{(q)},
\qquad
\mathbf{x}_{m}^{t+1}
\leftarrow
\tilde{\mathbf{x}}_{m}^{t+1,(q^{*})},
\label{eq:survival_selection}
\end{equation}
where \(\bar{c}_{m}^{(q)}\) denotes the average confidence of the tokens released in candidate block \(\tilde{\mathbf{x}}_{m}^{t+1,(q)}\) \textit{prior to mutation}.
This survival rule ensures that while mutation forces the exploration of structured numerical or symbolic alternatives, the final selection remains grounded in the model's natural distribution. This lightweight mechanism helps sampling-consistent failures escape high-confidence traps without disrupting the overall decoding stability.

\section{Experiment}
\label{sec:experiment}


\subsection{Implementation Details}

\mypara{Baselines and Metrics}
We compare Evolutionary Decoding with the default confidence-based decoding of LLaDA~2.0 and repeated sampling with pass@\emph{Num}. 
To analyze component contributions, we report \emph{Selection only} and the full \emph{Evolutionary Decoding} with structured mutation and survival. 
Since mutation and survival are designed to operate on selected candidate blocks, they are not treated as independent standalone decoders. 
We use pass@1 as the main metric and pass@8 to measure the gain achievable by repeated stochastic decoding. 
We also report trajectory-level diagnostics, including the average number of generated blocks, numerical/symbolic token ratio, repetition ratio, and confidence-collapse frequency, to examine whether decoding becomes reasoning-oriented.

\mypara{Datasets}
We evaluate on six mathematical reasoning benchmarks covering different difficulty levels and reasoning styles. 
AIME~2024/2025/2026 represent challenging competition-level problems requiring multi-step symbolic and numerical reasoning, while AMC~2023 contains shorter but still nontrivial contest problems. 
MATH500 provides broader coverage across algebra, geometry, counting, and so on, while GSM8K evaluates grade-school arithmetic reasoning in natural-language scenarios. 
Together, these datasets test whether Evolutionary Decoding improves both competition-style and general mathematical reasoning tasks. Details are provided in \textbf{Appendix~\ref{app:dataset_details}}.

\mypara{Setup}
All methods use frozen LLaDA~2.0 models without any parameter update, which are conducted with SGLang using KV-cache acceleration on 8 NVIDIA A100 GPUs. 
We keep the prompt format, maximum decoding length, block size, confidence threshold, and diffusion-step budget identical across methods. 
Unless otherwise stated, we use block size \(32\) and confidence threshold \(\tau=0.95\). 
For selection, we set \(\alpha=0.05\), \(\beta=0.2\), and \(K=3\). 
For mutation, we use \(\delta=0.2\), \(\tau_{\mathrm{mut}}=0.96\), and \(m_{\max}=16\). 
These hyperparameters are determined from the score formulations, with details in \textbf{Appendix~\ref{app:alpha_beta}}, and then verified by cross-comparison ablations on LLaDA2.0-Flash using AIME~2025. 
After selection, \textbf{the same configuration is directly transferred to all other benchmarks and to LLaDA2.0-mini}, without dataset-specific or model-specific tuning.

\begin{table*}[t]
\caption{
Main results of Evolutionary Decoding (ED) on six mathematical reasoning benchmarks.
Accuracy is reported as pass@1 accuracy (\%). 
Steps denotes average steps per problem (\textbf{Baseline: $\sim 10$ steps per block}).
Selection denotes the baseline with only selection without mutation.
\textbf{Hyperparameters are selected on Flash-AIME~2025 and then kept fixed across all experiments.}
}
\centering
\label{tab:main_results}
\resizebox{0.99\textwidth}{!}{
\begin{tabular}{ll cc cc cc cc cc cc}
\toprule[1pt]
\multirow{2}{*}{\textbf{Model}} 
& \multirow{2}{*}{\textbf{Decoding}}
& \multicolumn{2}{c}{\textbf{AIME24}}
& \multicolumn{2}{c}{\textbf{AIME25}}
& \multicolumn{2}{c}{\textbf{AIME26}}
& \multicolumn{2}{c}{\textbf{AMC23}}
& \multicolumn{2}{c}{\textbf{MATH500}}
& \multicolumn{2}{c}{\textbf{GSM8K}} \\
\cmidrule(lr){3-4}
\cmidrule(lr){5-6}
\cmidrule(lr){7-8}
\cmidrule(lr){9-10}
\cmidrule(lr){11-12}
\cmidrule(lr){13-14}
& 
& Acc. $\uparrow$ & Steps $\downarrow$
& Acc. $\uparrow$ & Steps $\downarrow$
& Acc. $\uparrow$ & Steps $\downarrow$
& Acc. $\uparrow$ & Steps $\downarrow$
& Acc. $\uparrow$ & Steps $\downarrow$
& Acc. $\uparrow$ & Steps $\downarrow$ \\
\midrule[1pt]
\multirow{3}{*}{LLaDA2.0-flash}
& Baseline
& 66.7 & 669.9
& 56.7 & 668.8
& 63.3 & 686.4
& 90.0 & 319.5
& 87.4 & 153.5
& 93.55 & 38.4 \\
& \textbf{Selection}
& 63.3 & 712.2
& 66.7 & 638.2
& \textbf{66.7} & 712.2
& 90.0 & 335.4
& 88.4 & 173.8
& \textbf{93.70} & 39.0 \\
& \textbf{ED}
& \textbf{70.0} & 960.4
& \textbf{70.0} & 1085.3
& \textbf{66.7} & 1112.7
& \textbf{95.0} & 878.3
& \textbf{88.8} & 310.4
& 93.17 & 69.5 \\
\midrule[1pt]
\multirow{3}{*}{LLaDA2.0-mini}
& Baseline
& 53.3 & 841.5
& 30.0 & 978.2
& 40.0 & 787.3
& 82.5 & 411.9
& 86.4 & 221.3
& 91.58 & 58.5 \\
& \textbf{Selection}
& 56.7 & 801.8
& 40.0 & 938.8
& \textbf{46.7} & 771.1
& 92.5 & 446.5
& 86.2 & 210.9
& \textbf{91.81} & 55.2 \\
& \textbf{ED}
& \textbf{60.0} & 1296.0
& \textbf{43.3} & 1320.6
& 40.0 & 1237.4
& \textbf{95.0} & 683.4
& \textbf{87.2} & 352.4
& 91.58 & 108.1 \\
\bottomrule[1pt]
\end{tabular}
}
\end{table*}

\begin{table*}[t]
\centering
\caption{
Comparison between standard decoding and Evolutionary Decoding under 8 test-time attempts (Temperature=0.7) with baseline @1.
Pass@8 is a metric indicating whether any of the 8 candidates is correct.
Majority@8 and BestConf@8 report practical final-answer selection results.
}
\label{tab:n8_comparison}
\resizebox{0.99\textwidth}{!}{
\begin{tabular}{llcccc cccc cccc}
\toprule
\multirow{2}{*}{\textbf{Model}} 
& \multirow{2}{*}{\textbf{Decoding}}
& \multicolumn{4}{c}{\textbf{AIME24}}
& \multicolumn{4}{c}{\textbf{AIME25}}
& \multicolumn{4}{c}{\textbf{AMC23}} \\
\cmidrule(lr){3-6}
\cmidrule(lr){7-10}
\cmidrule(lr){11-14}
& 
& P@1 & P@8 & Maj@8 & Best@8
& P@1 & P@8 & Maj@8 & Best@8
& P@1 & P@8 & Maj@8 & Best@8 \\
\midrule
\multirow{2}{*}{LLaDA2.0-flash}
& Baseline
& 66.7 & 73.3 & 60.0 & 66.7 
& 56.7 & 70.0 & 63.3 & 56.7 
& 90.0 & 92.5 & 82.5 & 90.0 \\
& \textbf{ED}
& 70.0 & 76.7 & 56.7 & 70.0
& 70.0 & 83.3 & 70.0 & 66.7 
& 95.0 & 95.0 & 85.0 & 95.0
\\
\midrule
\multirow{2}{*}{LLaDA2.0-mini}
& Baseline
& 53.3 & 66.7 & 40.0 & 53.3 
& 30.0 & 53.3 & 40.0 & 40.0 
& 82.5 & 95.0 & 92.5 & 82.5 \\
& \textbf{ED}
& 60.0 & 70.0 & 50.0 & 56.7 
& 43.3 & 60.0 & 43.3 & 46.7 
& 95.0 & 95.0 & 92.5 & 92.5 \\
\bottomrule
\end{tabular}
}
\end{table*}

\subsection{Main Results}

\mypara{Deterministic Trajectory Decoding}
We first evaluate Evolutionary Decoding in a deterministic setting, where each method produces one solution per problem. 
We directly \textit{transfer the hyperparameters selected on Flash-AIME25 to all other benchmarks without dataset-specific tuning}. 
As shown in Table~\ref{tab:main_results}, ED achieves consistent gains on most competition-style benchmarks, including AIME, AMC, and MATH500, indicating that the selection--mutation mechanism generalizes beyond the tuning dataset and stabilizes symbolic and numerical reasoning. 
Compared with Selection-only, full ED further improves several harder benchmarks, showing the value of structured block-wise mutation beyond step-wise selection. 
On GSM8K, where most problems require only about \textbf{4 to 5 blocks}, mutation brings limited benefit and may slightly hurt performance because short trajectories leave fewer steps for correcting perturbed candidates than selection. 
More GSM8K results are provided in \textbf{Appendix~\ref{app:gsm8k}}. 
Overall, ED trades more decoding steps for stronger reasoning trajectories, which is most beneficial on challenging benchmarks requiring longer symbolic reasoning.

\begin{table*}[t]
\centering
\caption{
Reference comparison with open-source dLLM reasoning methods.
PRISM~\cite{bai2026prism}, BACD+TCCF~\cite{lu2026advancingbacdtccf}, and $S^3$~\cite{bilal2026s} use LLaDA2.0-Mini, TDAR-8B-Thinking, and LLaDA-8B-Instruct as their respective baselines, while our ED on Mini from Table~\ref{tab:main_results}. Because protocols differ, we compare relative gains over each method's own baseline rather than absolute scores.}
\label{tab:opensource_reference}
\scriptsize
\setlength{\tabcolsep}{4pt}
\renewcommand{\arraystretch}{1.08}
\resizebox{0.8\textwidth}{!}{
\begin{tabular}{l|cc|cc|cc|cc}
\toprule
\textbf{Benchmark}
& \textbf{Baseline}
& \textbf{+ PRISM}
& \textbf{Baseline}
& \textbf{+ BACD+TCCF}
& \textbf{Baseline}
& \textbf{+ $S^3$}
& \textbf{Baseline}
& \textbf{+ ED} \\
\midrule
AIME24
& N/A & N/A
& 34.6 & 42.9
& N/A & N/A
& 53.3 & \textbf{60.0} \\
AIME25
& N/A & N/A
& 30.8 & 35.8
& N/A & N/A
& 30.0 & \textbf{43.3} \\
MATH500
& 30.60 & 32.60
& 81.60 & 84.00
& 25.60 & 30.20
& 86.4 & \textbf{87.2} \\
\bottomrule
\end{tabular}
}
\end{table*}

\mypara{Stochastic Perturbation Decoding}
We further evaluate ED under 8 stochastic perturbation attempts, as reported in Table~\ref{tab:n8_comparison}. 
This setting tests whether ED only improves a single deterministic trajectory, or whether its advantage also persists under temperature-based stochastic sampling. 
The results show that ED improves both direct decoding accuracy and the Pass@8 candidate-set upper bound compared with the Baseline. 
This indicates that ED does not simply refine one fixed trajectory, but helps overcome reasoning limitations that standard stochastic perturbations alone cannot reliably resolve. 
Importantly, Majority@8 and Best@8 are downstream selection strategies rather than competing decoding methods. 
Thus, ED is complementary to test-time scaling: it improves the quality of sampled candidates, while voting- or confidence-based selection can be applied.

\mypara{Comparison with other methods}
As shown in Table~\ref{tab:opensource_reference}, external test-time scaling methods improve their own baselines, but \textit{their absolute scores are not directly comparable} due to different evaluation protocols. 
PRISM improves LLaDA2.0-Mini on MATH500 by 2.00 points, $S^3$ improves LLaDA-8B-Instruct by 4.60 points, and BACD+TCCF improves TDAR-8B-Thinking by 8.3 points on AIME24/AIME25 and 2.40 points on MATH500. 
Under our same-protocol evaluation, ED improves LLaDA2.0-mini from 53.3 to 60.0 on AIME24, from 30.0 to 43.3 on AIME25, and from 86.4 to 87.2 on MATH500. 
Overall, ED achieves competitive gains among open-source scaling methods while using a diffusion trajectory and remaining compatible with parallel search or verification.

\subsection{Ablation Study}
\label{sub:ab}
Rather than treating the hyperparameters as purely empirical search variables, we first derive their reasonable ranges from the selection and mutation formulations, with details provided in \textbf{Appendix~\ref{app:alpha_beta}}. 
We then verify these choices through cross-comparison ablations on LLaDA2.0-Flash using AIME~2025, as shown in Tables~\ref{tab:ablation_selection} and~\ref{tab:ablation_mutation}. 
For selection, \(K=3\) performs best among different release budgets, suggesting that preserving a small set of promising numerical/symbolic tokens is more effective than overly conservative or aggressive release. 
Adding moderate entropy awareness further improves the average accuracy from \(60.0\%\) to \(66.7\%\), with the best setting \(\alpha=0.05\). 
For mutation, a moderate bias \(\delta=0.2\) works better than weaker or stronger perturbations, and the best performance is achieved when \(m_{\max}=16\) is combined with a slightly higher mutation threshold \(\tau_{\mathrm{mut}}=0.96\) or \(0.97\). 
This indicates that mutation is helpful for confident but trapped blocks, but premature or overly strong perturbation may destabilize decoding. 
After selecting this configuration, we directly transfer it to all other benchmarks without dataset-specific or model-specific tuning.


\begin{table*}[t]
\centering
\scriptsize
\setlength{\tabcolsep}{3pt}
\renewcommand{\arraystretch}{1.08}

\begin{minipage}[t]{0.45\textwidth}
\centering
\captionof{table}{
Selection ablation on Flash-AIME 2025.
All settings use block size 32 and confidence threshold $\tau=0.95$ with Pass@1.
$\alpha$ and $\beta$ denote the selection weights with $K$. 
}
\label{tab:ablation_selection}

\resizebox{0.9\linewidth}{!}{
\begin{tabular}{lccc}
\toprule
\textbf{Configuration} 
& \textbf{AIME-I} 
& \textbf{AIME-II} 
& \textbf{Avg.} \\
\midrule
$\alpha{=}0.00$, $\beta{=}0.2$, $K{=}1$
& 46.7 & 60.0 & 53.3 \\
$\alpha{=}0.00$, $\beta{=}0.2$, $K{=}2$
& 46.7 & 66.7 & 56.7 \\
$\alpha{=}0.00$, $\beta{=}0.2$, $K{=}3$
& 60.0 & 60.0 & 60.0 \\

$\alpha{=}0.00$, $\beta{=}0.2$, $K{=}4$
& 53.3 & 60.0 & 56.7 \\

$\alpha{=}0.00$, $\beta{=}0.2$, $K{=}5$
& 46.7 & 60.0 & 53.3 \\

$\alpha{=}0.08$, $\beta{=}0.2$, $K{=}3$
& 53.3 & 66.7 & 60.0 \\

$\alpha{=}0.07$, $\beta{=}0.2$, $K{=}3$
& 60.0 & 60.0 & 60.0 \\

$\alpha{=}0.05$, $\beta{=}0.2$, $K{=}3$
& \textbf{60.0} & \textbf{73.3} & \textbf{66.7} \\
\bottomrule
\end{tabular}
}

\end{minipage}
\hfill
\begin{minipage}[t]{0.525\textwidth}
\centering
\captionof{table}{
Mutation ablation on Flash-AIME 2025.
All settings are built on Selection mechanism.
$\delta$, $\tau_{\mathrm{mut}}$,  and $m_{\max}$ denote the mutation bias, threshold, and the maximum number of mutated blocks.
}
\label{tab:ablation_mutation}

\resizebox{0.9\linewidth}{!}{
\begin{tabular}{lccc}
\toprule
\textbf{Configuration} 
& \textbf{AIME-I} 
& \textbf{AIME-II} 
& \textbf{Avg.} \\
\midrule

$\delta{=}0.1$, $\tau_{\mathrm{mut}}{=}0.95$, $m_{\max}{=}16$
& 66.7 & 53.3 & 60.0 \\

$\delta{=}0.2$, $\tau_{\mathrm{mut}}{=}0.95$, $m_{\max}{=}8$
& 60.0 & 73.3 & 66.7 \\

$\delta{=}0.2$, $\tau_{\mathrm{mut}}{=}0.95$, $m_{\max}{=}16$
& 46.7 & 73.3 & 60.0 \\

$\delta{=}0.2$, $\tau_{\mathrm{mut}}{=}0.96$, $m_{\max}{=}8$
& 66.7 & 66.7 & 66.7 \\

$\delta{=}0.2$, $\tau_{\mathrm{mut}}{=}0.96$, $m_{\max}{=}16$
& \textbf{66.7} & \textbf{73.3} & \textbf{70.0} \\

$\delta{=}0.2$, $\tau_{\mathrm{mut}}{=}0.97$, $m_{\max}{=}16$
& \underline{66.7} & \underline{73.3} & \underline{70.0} \\

$\delta{=}0.3$, $\tau_{\mathrm{mut}}{=}0.99$, $m_{\max}{=}8$
& 66.7 & 66.7 & 66.7 \\

$\delta{=}0.3$, $\tau_{\mathrm{mut}}{=}0.99$, $m_{\max}{=}16$
& 60.0 & 66.7 & 63.3 \\

\bottomrule
\end{tabular}
}

\end{minipage}

\end{table*}

\subsection{Analysis}

\begin{figure*}[t]
    \centering

    \begin{subfigure}[t]{0.49\textwidth}
        \centering
        \includegraphics[width=\linewidth]{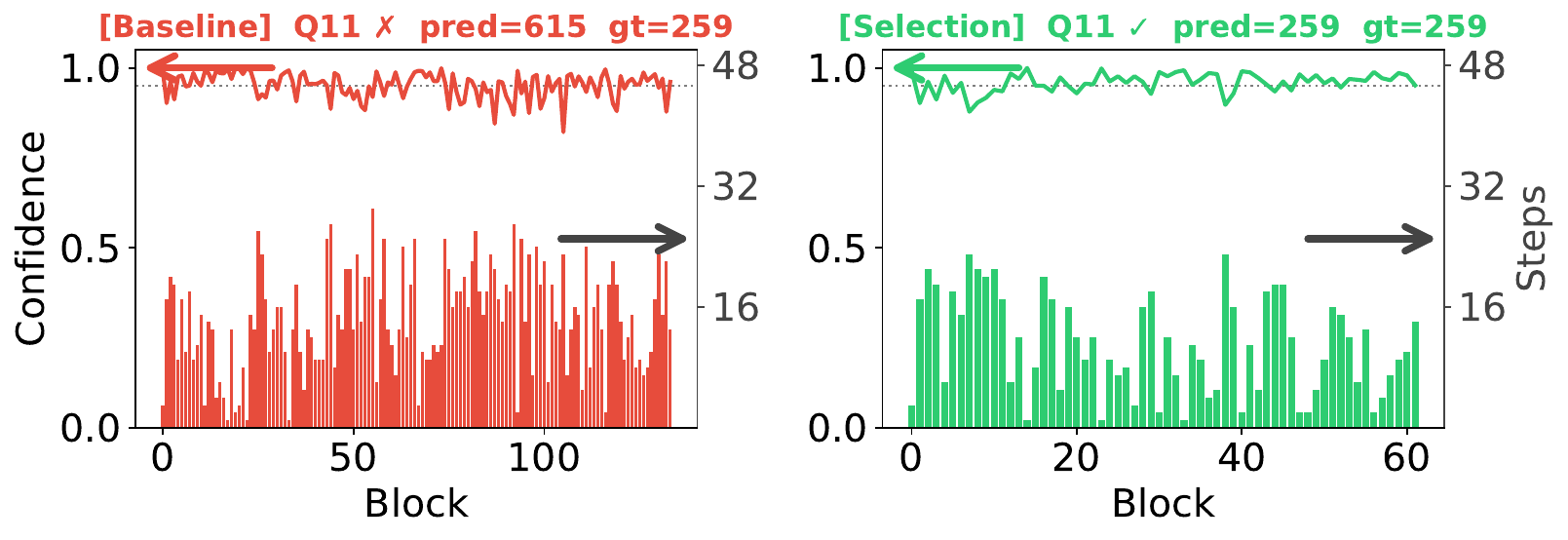}
        \put(-70,50){{\fontsize{5}{6}\selectfont\bfseries Selection}}
        \caption{AIME-I-11: Selection improves Baseline.}
        \label{fig:aime_i05_selection_baseline}
    \end{subfigure}
    \hfill
    \begin{subfigure}[t]{0.49\textwidth}
        \centering
        \includegraphics[width=\linewidth]{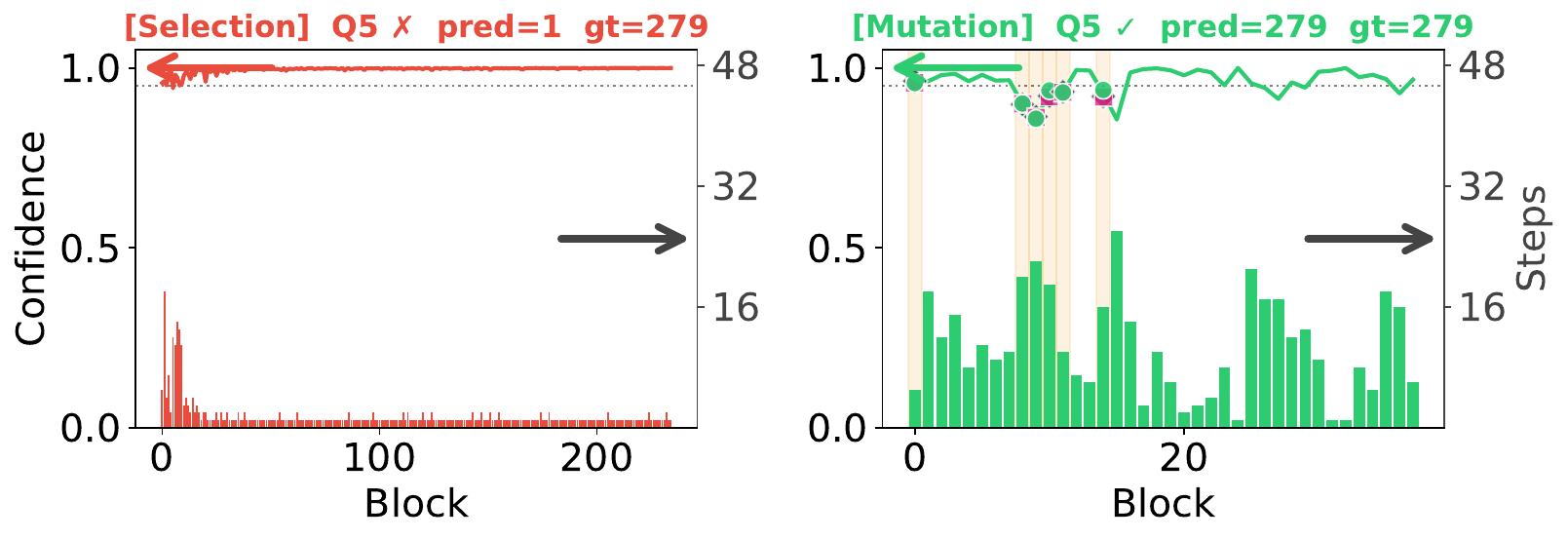}
        \put(-80,45){{\fontsize{5}{6}\selectfont\bfseries Mutation}}
        \caption{AIME-I-05: Mutation improves Selection.}
        \label{fig:aime_i05_mutation_selection}
    \end{subfigure}

    \caption{
    Case studies on AIME-I-05 showing how Selection and Mutation further works. The full answer are provided in the \textbf{Supplementary Material} and more cases in \textbf{Figure~\ref{fig:table_aime}-\ref{fig:m_aime2}}. 
    }
    \label{fig:aime_i05_case_study}
\end{figure*}

\mypara{Case Analysis}
Figure~\ref{fig:aime_i05_case_study} illustrates how Selection and Mutation address different failure modes on AIME-I-05/11. 
Selection corrects the token-release process by prioritizing numerical and symbolic tokens over locally confident, guiding the trajectory toward the answer-bearing reasoning path. 
When the selected trajectory still falls into a confident but incorrect direction, Mutation introduces structured numerical-symbolic perturbations in early blocks to create alternative branches and escape the local reasoning basin. 
Overall, this case shows that ED mitigates the confidence trap through two complementary mechanisms: Selection preserves useful reasoning signals, while Mutation provides structured exploration when the trajectory remains trapped in high-confidence.

\begin{wraptable}{r}{0.43\linewidth}
\centering
\caption{Gated mutation analysis.}
\label{tab:gated_mutation}
\setlength{\tabcolsep}{4pt}
\renewcommand{\arraystretch}{1.05}
\begin{tabular}{lccc}
\toprule
\textbf{Setting} & \textbf{Blocks} & \textbf{Steps} & \textbf{Solved} \\
\midrule
Selection & 54.9 & 638  & 66.7\% \\
Mutation  & 63.1 & 1085 & 70.0\% \\
Gated     & 50.6 & 1230 & 73.3\% \\
\bottomrule
\end{tabular}
\end{wraptable}
\textbf{Gated Mutation.}
Mutation helps recover hard cases by exploring alternative branches, but applying it to all problems can be unnecessary or even harmful, since it may add extra steps and disturb trajectories that Selection already solves correctly.
We therefore consider a gated mutation strategy, where mutation is triggered only when the Selection trajectory is likely to fail, e.g., \(b>64\) with confidence below \(\tau_{\mathrm{mut}}\), or last-10-block repetition above \(0.5\). 
As shown in Table~\ref{tab:gated_mutation}, Gated Mutation improves the solved count to \(73.3\%\) with little step cost compared with unconditional Mutation, suggesting that mutation is useful as a failure-aware correction after Selection.
Details are provided in \textbf{Appendix~\ref{app:tra_amie}}, including \textbf{Figures~\ref{fig:table_aime}-\ref{fig:m_aime2}}.

\section{Conclusion}
\label{sec:conclusion}

This work highlights a key bottleneck in diffusion LLM reasoning: failures are not always due to insufficient sampling, but can arise from the internal dynamics of confidence-driven block-wise decoding. 
Our trajectory analysis shows that repeated sampling often reproduces the same unstable or incorrect reasoning patterns, revealing that local confidence can misguide the formation of numerical-symbolic reasoning. 
This motivates a shift from output-level test-time scaling to trajectory-level intervention.
We proposed Evolutionary Decoding, a training-free framework that reshapes the evolution of reasoning states during decoding. 
By preserving fragile numerical-symbolic signals and introducing structured alternatives before a block collapses into an incorrect direction, ED improves both deterministic decoding and stochastic multi-sample candidate quality. 
These results demonstrate that reliable reasoning in diffusion LLMs requires not only more samples, but better control over how reasoning trajectories are formed, selected, and diversified.

\bibliographystyle{ieeetr}
\bibliography{reference}

@article{austin2021structured,
  title={Structured denoising diffusion models in discrete state-spaces},
  author={Austin, Jacob and Johnson, Daniel D and Ho, Jonathan and Tarlow, Daniel and Van Den Berg, Rianne},
  journal={Advances in neural information processing systems},
  volume={34},
  pages={17981--17993},
  year={2021}
}

@article{gulrajani2023likelihood,
  title={Likelihood-based diffusion language models},
  author={Gulrajani, Ishaan and Hashimoto, Tatsunori B},
  journal={Advances in Neural Information Processing Systems},
  volume={36},
  pages={16693--16715},
  year={2023}
}

@article{lou2023discrete,
  title={Discrete diffusion modeling by estimating the ratios of the data distribution},
  author={Lou, Aaron and Meng, Chenlin and Ermon, Stefano},
  journal={arXiv preprint arXiv:2310.16834},
  year={2023}
}

@article{shi2024simplified,
  title={Simplified and generalized masked diffusion for discrete data},
  author={Shi, Jiaxin and Han, Kehang and Wang, Zhe and Doucet, Arnaud and Titsias, Michalis},
  journal={Advances in neural information processing systems},
  volume={37},
  pages={103131--103167},
  year={2024}
}

@article{sahoo2024simple,
  title={Simple and effective masked diffusion language models},
  author={Sahoo, Subham S and Arriola, Marianne and Schiff, Yair and Gokaslan, Aaron and Marroquin, Edgar and Chiu, Justin T and Rush, Alexander and Kuleshov, Volodymyr},
  journal={Advances in Neural Information Processing Systems},
  volume={37},
  pages={130136--130184},
  year={2024}
}

@article{nie2024scaling,
  title={Scaling up masked diffusion models on text},
  author={Nie, Shen and Zhu, Fengqi and Du, Chao and Pang, Tianyu and Liu, Qian and Zeng, Guangtao and Lin, Min and Li, Chongxuan},
  journal={arXiv preprint arXiv:2410.18514},
  year={2024}
}

@article{arriola2025block,
  title={Block diffusion: Interpolating between autoregressive and diffusion language models},
  author={Arriola, Marianne and Gokaslan, Aaron and Chiu, Justin T and Yang, Zhihan and Qi, Zhixuan and Han, Jiaqi and Sahoo, Subham Sekhar and Kuleshov, Volodymyr},
  journal={arXiv preprint arXiv:2503.09573},
  year={2025}
}

@article{nie2025large,
  title={Large language diffusion models},
  author={Nie, Shen and Zhu, Fengqi and You, Zebin and Zhang, Xiaolu and Ou, Jingyang and Hu, Jun and Zhou, Jun and Lin, Yankai and Wen, Ji-Rong and Li, Chongxuan},
  journal={arXiv preprint arXiv:2502.09992},
  year={2025}
}

@article{bie2025llada2,
  title={Llada2. 0: Scaling up diffusion language models to 100b},
  author={Bie, Tiwei and Cao, Maosong and Chen, Kun and Du, Lun and Gong, Mingliang and Gong, Zhuochen and Gu, Yanmei and Hu, Jiaqi and Huang, Zenan and Lan, Zhenzhong and others},
  journal={arXiv preprint arXiv:2512.15745},
  year={2025}
}

@article{zhu2025llada,
  title={Llada 1.5: Variance-reduced preference optimization for large language diffusion models},
  author={Zhu, Fengqi and Wang, Rongzhen and Nie, Shen and Zhang, Xiaolu and Wu, Chunwei and Hu, Jun and Zhou, Jun and Chen, Jianfei and Lin, Yankai and Wen, Ji-Rong and others},
  journal={arXiv preprint arXiv:2505.19223},
  year={2025}
}

@article{khanna2025mercury,
  title={Mercury: Ultra-fast language models based on diffusion},
  author={Khanna, Samar and Kharbanda, Siddhant and Li, Shufan and Varma, Harshit and Wang, Eric and Birnbaum, Sawyer and Luo, Ziyang and Miraoui, Yanis and Palrecha, Akash and Ermon, Stefano and others},
  journal={arXiv e-prints},
  pages={arXiv--2506},
  year={2025}
}

@article{ye2025dream,
  title={Dream 7b: Diffusion large language models},
  author={Ye, Jiacheng and Xie, Zhihui and Zheng, Lin and Gao, Jiahui and Wu, Zirui and Jiang, Xin and Li, Zhenguo and Kong, Lingpeng},
  journal={arXiv preprint arXiv:2508.15487},
  year={2025}
}

@article{zhu2025lladamoe,
  title={Llada-moe: A sparse moe diffusion language model},
  author={Zhu, Fengqi and You, Zebin and Xing, Yipeng and Huang, Zenan and Liu, Lin and Zhuang, Yihong and Lu, Guoshan and Wang, Kangyu and Wang, Xudong and Wei, Lanning and others},
  journal={arXiv preprint arXiv:2509.24389},
  year={2025}
}

@article{wei2022chain,
  title={Chain-of-thought prompting elicits reasoning in large language models},
  author={Wei, Jason and Wang, Xuezhi and Schuurmans, Dale and Bosma, Maarten and Xia, Fei and Chi, Ed and Le, Quoc V and Zhou, Denny and others},
  journal={Advances in neural information processing systems},
  volume={35},
  pages={24824--24837},
  year={2022}
}

@article{wang2022self,
  title={Self-consistency improves chain of thought reasoning in language models},
  author={Wang, Xuezhi and Wei, Jason and Schuurmans, Dale and Le, Quoc and Chi, Ed and Narang, Sharan and Chowdhery, Aakanksha and Zhou, Denny},
  journal={arXiv preprint arXiv:2203.11171},
  year={2022}
}

@article{cobbe2021training,
  title={Training verifiers to solve math word problems},
  author={Cobbe, Karl and Kosaraju, Vineet and Bavarian, Mohammad and Chen, Mark and Jun, Heewoo and Kaiser, Lukasz and Plappert, Matthias and Tworek, Jerry and Hilton, Jacob and Nakano, Reiichiro and others},
  journal={arXiv preprint arXiv:2110.14168},
  year={2021}
}

@article{snell2024scaling,
  title={Scaling llm test-time compute optimally can be more effective than scaling model parameters},
  author={Snell, Charlie and Lee, Jaehoon and Xu, Kelvin and Kumar, Aviral},
  journal={arXiv preprint arXiv:2408.03314},
  year={2024}
}

@article{ye2024diffusion,
  title={Diffusion of thought: Chain-of-thought reasoning in diffusion language models},
  author={Ye, Jiacheng and Gong, Shansan and Chen, Liheng and Zheng, Lin and Gao, Jiahui and Shi, Han and Wu, Chuan and Jiang, Xin and Li, Zhenguo and Bi, Wei and others},
  journal={Advances in Neural Information Processing Systems},
  volume={37},
  pages={105345--105374},
  year={2024}
}

@article{ye2024beyond,
  title={Beyond autoregression: Discrete diffusion for complex reasoning and planning},
  author={Ye, Jiacheng and Gao, Jiahui and Gong, Shansan and Zheng, Lin and Jiang, Xin and Li, Zhenguo and Kong, Lingpeng},
  journal={arXiv preprint arXiv:2410.14157},
  year={2024}
}

@article{zhao2025d1,
  title={d1: Scaling reasoning in diffusion large language models via reinforcement learning},
  author={Zhao, Siyan and Gupta, Devaansh and Zheng, Qinqing and Grover, Aditya},
  journal={arXiv preprint arXiv:2504.12216},
  year={2025}
}

@article{zhu2025dirl,
  title={Dirl: An efficient post-training framework for diffusion language models},
  author={Zhu, Ying and Wan, Jiaxin and Liu, Xiaoran and He, Siyang and Wang, Qiqi and Guo, Xu and Liang, Tianyi and Huang, Zengfeng and He, Ziwei and Qiu, Xipeng},
  journal={arXiv preprint arXiv:2512.22234},
  year={2025}
}

@article{chen2025reasoning,
  title={Reasoning in Diffusion Large Language Models is Concentrated in Dynamic Confusion Zones},
  author={Chen, Ranfei and Chen, Ming and Wang, Kaifei},
  journal={arXiv preprint arXiv:2511.15208},
  year={2025}
}

@article{shen2026improving,
  title={Improving Diffusion Language Model Decoding through Joint Search in Generation Order and Token Space},
  author={Shen, Yangyi and Feng, Tianjian and Han, Jiaqi and Wang, Wen and Chen, Tianlang and Shen, Chunhua and Leskovec, Jure and Ermon, Stefano},
  journal={arXiv preprint arXiv:2601.20339},
  year={2026}
}

@article{cai2026confidence,
  title={Confidence-Based Decoding is Provably Efficient for Diffusion Language Models},
  author={Cai, Changxiao and Li, Gen},
  journal={arXiv preprint arXiv:2603.22248},
  year={2026}
}

@article{fang2026locally,
  title={Locally Confident, Globally Stuck: The Quality-Exploration Dilemma in Diffusion Language Models},
  author={Fang, Liancheng and Liu, Aiwei and Zou, Henry Peng and Chen, Yankai and Ma, Enze and Pan, Leyi and Miao, Chunyu and Huang, Wei-Chieh and Liu, Xue and Yu, Philip S},
  journal={arXiv preprint arXiv:2604.00375},
  year={2026}
}

@article{bilal2026s,
  title={S3: Stratified Scaling Search for Test-Time in Diffusion Language Models},
  author={Bilal, Ahsan and Mohsin, Muhammad Ahmed and Umer, Muhammad and Aali, Asad and Khanzada, Muhammad Usman and Rafique, Muhammad Usman and He, Zihao and Fox, Emily and Hougen, Dean F},
  journal={arXiv preprint arXiv:2604.06260},
  year={2026}
}

@article{chen2025rfg,
  title={Rfg: Test-time scaling for diffusion large language model reasoning with reward-free guidance},
  author={Chen, Tianlang and Xu, Minkai and Leskovec, Jure and Ermon, Stefano},
  journal={arXiv preprint arXiv:2509.25604},
  year={2025}
}

@article{wang2503remasking,
  title={Remasking discrete diffusion models with inference-time scaling (2025)},
  author={Wang, Guanghan and Schiff, Yair and Sahoo, Subham Sekhar and Kuleshov, Volodymyr},
  journal={arXiv preprint arXiv:2503.00307}
}

@article{kang2025scalablebestN,
  title={Scalable best-of-n selection for large language models via self-certainty},
  author={Kang, Zhewei and Zhao, Xuandong and Song, Dawn},
  journal={arXiv preprint arXiv:2502.18581},
  year={2025}
}

@article{lee2025testmath,
  title={Test-Time Scaling in Diffusion LLMs via Hidden Semi-Autoregressive Experts},
  author={Lee, Jihoon and Moon, Hoyeon and Zhai, Kevin and Chithanar, Arun Kumar and Sahu, Anit Kumar and Kar, Soummya and Lee, Chul and Chakraborty, Souradip and Bedi, Amrit Singh},
  journal={arXiv preprint arXiv:2510.05040},
  year={2025}
}

@article{lu2026advancingbacdtccf,
  title={Advancing Block Diffusion Language Models for Test-Time Scaling},
  author={Lu, Yi and Kong, Deyang and Wang, Jianing and Guo, Linsen and Wang, Xue and Guo, Qi and Gui, Tao and Huang, Xuanjing and Ye, Wei and Zhang, Shikun and others},
  journal={arXiv preprint arXiv:2602.09555},
  year={2026}
}

@article{bai2026prism,
  title={Prism: Efficient Test-Time Scaling via Hierarchical Search and Self-Verification for Discrete Diffusion Language Models},
  author={Bai, Jinbin and Li, Yixuan and Zhu, Yuchen and Xin, Yi and Shi, Qingyu and Feng, Aosong and Liu, Xiaohong and Tao, Molei and Xue, Jianru and Li, Xiangtai and others},
  journal={arXiv preprint arXiv:2602.01842},
  year={2026}
}

@misc{eiben2003introduction,
  title={Introduction to Evolutionary Computing},
  author={Eiben, AE and Smith, JE},
  year={2003},
  publisher={Springer}
}

@inproceedings{de2017evolutionary,
  title={Evolutionary computation: a unified approach},
  author={De Jong, Kenneth},
  booktitle={Proceedings of the Genetic and Evolutionary Computation Conference Companion},
  pages={373--388},
  year={2017}
}

@article{guo2023connecting,
  title={Connecting large language models with evolutionary algorithms yields powerful prompt optimizers},
  author={Guo, Qingyan and Wang, Rui and Guo, Junliang and Li, Bei and Song, Kaitao and Tan, Xu and Liu, Guoqing and Bian, Jiang and Yang, Yujiu},
  journal={arXiv preprint arXiv:2309.08532},
  year={2023}
}

@article{chao2024large,
  title={When large language models meet evolutionary algorithms},
  author={Chao, Wang and Zhao, Jiaxuan and Jiao, Licheng and Li, Lingling and Liu, Fang and Yang, Shuyuan},
  journal={arXiv e-prints},
  pages={arXiv--2401},
  year={2024}
}

@article{zheng2024sglang,
  title={Sglang: Efficient execution of structured language model programs},
  author={Zheng, Lianmin and Yin, Liangsheng and Xie, Zhiqiang and Sun, Chuyue and Huang, Jeff and Yu, Cody H and Cao, Shiyi and Kozyrakis, Christos and Stoica, Ion and Gonzalez, Joseph E and others},
  journal={Advances in neural information processing systems},
  volume={37},
  pages={62557--62583},
  year={2024}
}


\appendix

\newpage
\section{Implementation Details}
\label{app:imp}

\subsection{Dataset Details}
\label{app:dataset_details}

We provide additional details of the evaluation datasets in Table~\ref{tab:dataset_details}. 
The benchmarks cover different levels of mathematical reasoning, ranging from grade-school arithmetic word problems to challenging competition-style problems. 
AIME-style datasets mainly require multi-step symbolic reasoning and exact numerical answers, while AMC contains shorter contest problems. 
MATH500 provides broader topic coverage across high-school and competition mathematics, and GSM8K evaluates natural-language arithmetic reasoning.

\begin{table*}[t]
\centering
\caption{
Details of the mathematical reasoning benchmarks used in our evaluation.
}
\label{tab:dataset_details}
\vspace{5pt}
\resizebox{\linewidth}{!}{
\begin{tabular}{lcll}
\toprule
\textbf{Dataset} & \textbf{\#Problems} & \textbf{Difficulty / Type} & \textbf{Answer Format} \\
\midrule
AIME~2024 
& 30 
& High-school competition; multi-step symbolic and numerical reasoning 
& Integer answer \\

AIME~2025 
& 30 
& High-school competition; challenging symbolic reasoning 
& Integer answer \\

AIME~2026 
& 30 
& High-school competition; challenging symbolic and numerical reasoning 
& Integer answer \\

AMC~2023 
& 40 
& Middle/high-school contest problems; relatively shorter reasoning chains 
& Reference answer \\

MATH500 
& 500 
& Diverse competition-level mathematics across algebra, geometry, counting, and etc. 
& Short mathematical answer \\

GSM8K 
& 1319 
& Grade-school arithmetic word problems with natural-language reasoning 
& Numerical answer \\
\bottomrule
\end{tabular}
}
\end{table*}

\subsection{Prompt and Answer Extraction}
\label{app:prompt_answer_extraction}

For all benchmarks, we use the same simple system prompt without dataset-specific prompt engineering:
\begin{quote}
\texttt{Please reason step by step, and put your final answer within \textbackslash boxed\{\}.}
\end{quote}

During evaluation, we extract the final answer by matching the content inside \(\backslash\mathrm{boxed}\{\cdot\}\). 
If multiple boxed expressions appear, we use the final matched expression as the model prediction. 
The extracted answer is then normalized by removing redundant spaces, line breaks, and simple formatting differences before being compared with the reference answer. 
This rule is applied consistently to all methods, models, and datasets.

We note that this regex-based answer extraction may not exactly match the evaluation scripts used in some external reports, especially when a model produces unusual formatting, multiple candidate answers, or equivalent mathematical expressions in different forms. 
However, within our experiments, all baselines and proposed variants are evaluated using the same prompt, the same decoding constraints, and the same answer extraction pipeline. 
Therefore, the reported comparisons are internally consistent and reflect the relative effect of the proposed decoding strategy under an identical evaluation protocol.

\subsection{Implementation of Numerical and Symbolic Token Detection}
In our implementation, numerical and symbolic tokens are identified by a simple lexical matching rule. 
Specifically, after decoding each candidate token into text, we use regular expressions to check whether the token consists only of digits, only predefined mathematical symbols, or their combination:
\begin{quote}
\small
\begin{verbatim}
_DIGIT_RE  = re.compile(r'^\d+$')
_SYMBOL_RE = re.compile(r'^[\+\-\*\/\=\(\)\[\]\{\}\,\.\:\;\\]+$')
_DIGIT_SYMBOL_RE = re.compile(r'^[\d\+\-\*\/\=\(\)\[\]\{\}\,\.\:\;\\]+$')
\end{verbatim}
\end{quote}

Here, \texttt{\_DIGIT\_RE} detects pure numerical tokens, \texttt{\_SYMBOL\_RE} detects tokens composed only of the predefined symbolic characters, and \texttt{\_DIGIT\_SYMBOL\_RE} detects tokens containing only digits and these symbols. 
This rule is intentionally simple and tokenizer-agnostic, serving as a lightweight proxy for identifying tokens that are likely to be important in mathematical reasoning. 
However, it does not cover all possible mathematical expressions, such as alphabetic variables, inequalities, powers, percentages, or task-specific operators. 
A richer symbol vocabulary or tokenizer-aware mathematical token classifier could potentially capture more reasoning-relevant tokens and further improve the effectiveness of selection and mutation.

\section{Analysis on Main Results}
\label{app:main}

\begin{figure}
    \centering
    \includegraphics[width=1\linewidth]{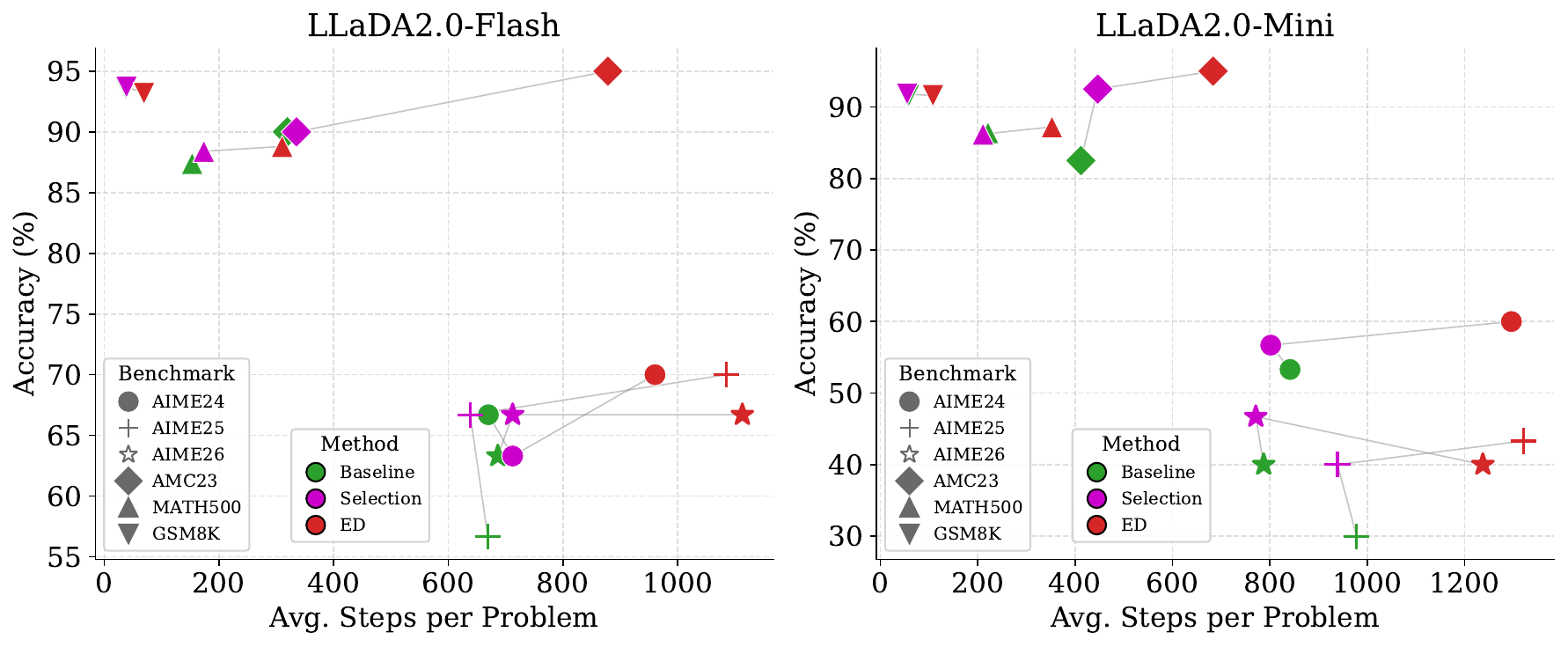}
    \caption{Steps vs. Accuracy trade-off of Baseline, Selection, and ED across six benchmarks. Each point represents one benchmark evaluated under one decoding method. Grey lines connect the same benchmark across the three methods (Baseline → Selection → ED), illustrating the trajectory in the steps–accuracy space. Left: LLaDA2.0-Flash. Right: LLaDA2.0-Mini.}
    \label{fig:steps_acc}
\end{figure}

\mypara{Steps vs. Accuracy for Table~\ref{tab:main_results}}
Figure~\ref{fig:steps_acc} further analyzes the computation--accuracy trade-off of different decoding strategies. 
Across both LLaDA2.0-Flash and LLaDA2.0-Mini, Selection stays close to the Baseline in the steps--accuracy space, suggesting that step-wise token selection introduces little extra decoding cost but also provides limited accuracy improvement. 
In contrast, ED moves most benchmark points substantially to the right, indicating a clear increase in decoding steps, but whether this additional computation is worthwhile depends on the benchmark difficulty. 
For challenging competition-style datasets such as AIME24/25/26 and AMC23, the trajectories generally move upward together with the step increase, showing that ED can convert additional test-time computation into better reasoning accuracy. 
This trend is especially visible on AMC23, where ED improves both models to 95.0\%, and on the AIME benchmarks, where the gains are more pronounced than those from Selection alone. 
For MATH500, the improvement is more moderate, suggesting that ED remains helpful but with a smaller marginal return. 
For GSM8K, where most problems require only short reasoning trajectories with about \textbf{4 to 5 blocks}, mutation does not provide clear gains and may slightly hurt performance compared to selection. 
This is likely because short trajectories leave insufficient decoding steps for mutated candidates to be corrected and stabilized, making mutation-induced perturbations not useful.
Overall, the figure shows that ED is not simply a uniformly more expensive decoder; rather, its additional computation is most valuable on harder reasoning benchmarks where escaping local decoding traps for meaningful accuracy gains.

\begin{figure}
    \centering
    \includegraphics[width=1\linewidth]{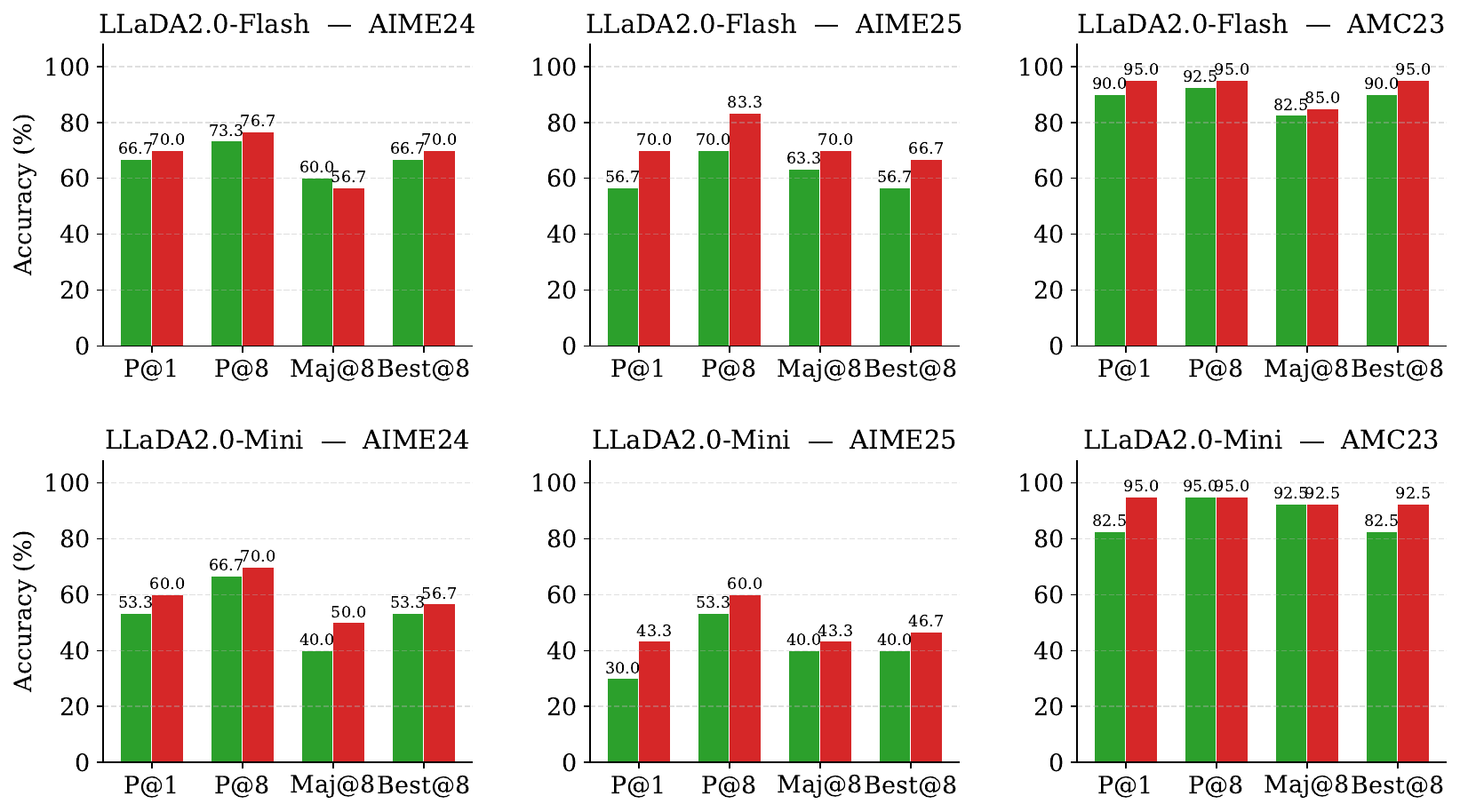}
    \caption{Comparison of Baseline and ED under 8 test-time attempts across four metrics. Red denotes ED while Green denotes baseline.}
    \label{fig:n8_compare}
\end{figure}

\section{Analysis on Multiple Sampling}
\label{app:tra_multi}

\mypara{Comparison of Baseline and ED for Table~\ref{tab:n8_comparison}} 
Figure~\ref{fig:n8_compare} compares Baseline and ED in the multi-sample regime with 8 times sampling. 
ED improves both P@1 and P@8 across the two models, showing that its structured exploration benefits not only a single decoding trajectory but also the overall candidate pool. 
However, the large gap between P@8 and the practical selection metrics indicates that generating a correct candidate is often easier than selecting it. 
Majority voting may suppress minority-correct trajectories, while confidence-based selection remains limited by imperfect confidence calibration. 
This issue is more evident on challenging AIME problems, whereas AMC23 shows more consistent candidate agreement under ED. 
Therefore, the results suggest that ED already improves multi-sample exploration, but its full potential depends on stronger final-answer selection strategies that can better approach the P@8 oracle.

\section{Hyperparameter Calibration for Step-wise Selection}
\label{app:alpha_beta}

This section explains how the two selection hyperparameters, \(\alpha\) and \(\beta\), are determined from the release boundary rather than by exhaustive grid search. 
In confidence-based diffusion decoding, the release threshold is typically set to a high value, e.g., \(\tau=0.95\), so that only sufficiently confident tokens are unmasked. 
While this conservative rule improves reliability, it may delay reasoning-critical numerical or symbolic tokens whose confidence is slightly below the threshold, and may also release repetitive tokens when the block-level continuation becomes over-confident. 
Therefore, we calibrate \(\alpha\) and \(\beta\) around this high-confidence boundary: \(\alpha\) specifies how much uncertainty is allowed to promote numerical-symbolic tokens near \(\tau\), while \(\beta\) specifies how much repetition is needed to suppress borderline releases.

Recall that, for a masked position \(i\) in block \(m\) at step \(t\), the release score is
\begin{equation}
s_{m,i}^{t}
=
c_{m,i}^{t}
+
\alpha \eta_{m,i}^{t} H_{m,i}^{t}
-
\beta r_m^t,
\end{equation}
and the token is released when \(s_{m,i}^{t}\geq \tau\). 
Here, \(c_{m,i}^{t}\) is the top-1 confidence, \(H_{m,i}^{t}\) is the token entropy, \(\eta_{m,i}^{t}\in\{0,1\}\) indicates whether the top-1 prediction is a numerical or symbolic token, \(r_m^t\) is the block-level repetition ratio, and \(\tau\) is the release threshold.

\begin{figure}[t]
    \centering
    \begin{subfigure}[b]{0.49\linewidth}
        \centering
        \includegraphics[width=\linewidth]{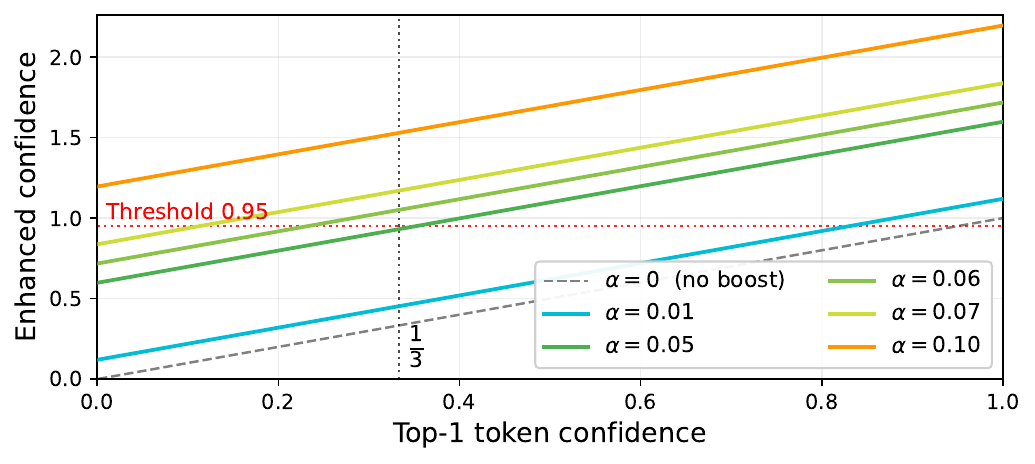}
        \caption{Simulation on loose formulation of Eq.~(\ref{eq:alpha_loose}).}
        \label{fig:en_wight_a}
    \end{subfigure}
    \hfill
    \begin{subfigure}[b]{0.49\linewidth}
        \centering
        \includegraphics[width=\linewidth]{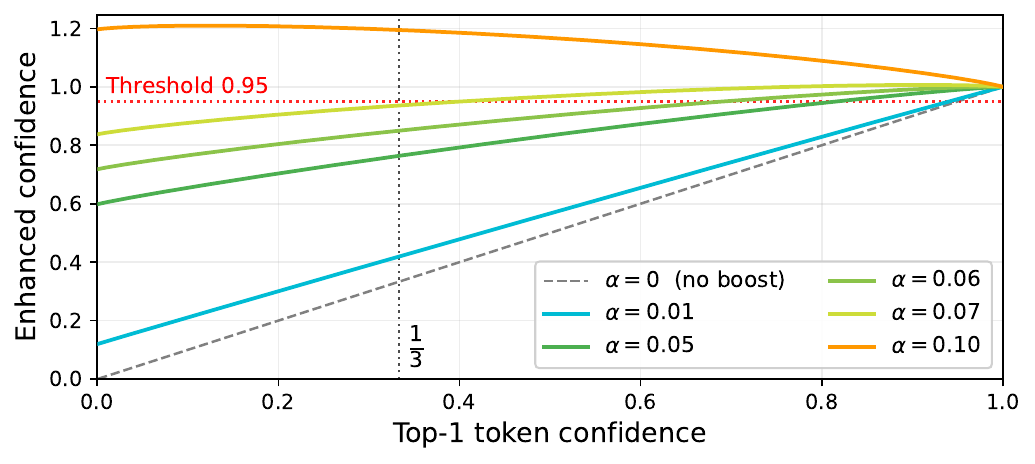}
        \caption{Simulation on strict formulation of Eq.(~\ref{eq:alpha_strict}).}
        \label{fig:en_wight_b}
    \end{subfigure}
    \caption{The relationship between the Top-1 confidence and weight for entropy enhancement of digit/symbol tokens in the process of step-wise selection.}
    \label{fig:entropy_weight_analysis}
\end{figure}

\mypara{Calibration of the entropy enhancement weight \(\alpha\)}
The role of \(\alpha\) is to allow a reasoning-critical numerical or symbolic token to be released even when its confidence is slightly below the high release threshold, e.g., \(\tau=0.95\).
We therefore calibrate \(\alpha\) by specifying a target confidence \(c_{\mathrm{tar}}<\tau\): a numerical or symbolic token with confidence \(c_{\mathrm{tar}}\) should just reach the release boundary when its predictive distribution still contains sufficient uncertainty. 
Ignoring repetition first, the boundary condition is
\begin{equation}
c_{\mathrm{tar}}+\alpha H_{\mathrm{ref}}=\tau,
\label{eq:alpha_boundary}
\end{equation}
where \(H_{\mathrm{ref}}\) denotes a reference entropy level. 
This gives
\begin{equation}
\alpha
=
\frac{\tau-c_{\mathrm{tar}}}{H_{\mathrm{ref}}}.
\label{eq:alpha_general}
\end{equation}

A loose calibration uses the global entropy upper bound \(H\leq \log V\), where \(V\) is the vocabulary size. 
This gives
\begin{equation}
\alpha_{\mathrm{loose}}
=
\frac{\tau-c_{\mathrm{tar}}}{\log V}.
\label{eq:alpha_loose}
\end{equation}
This estimate is optimistic because it assumes that the maximum possible vocabulary entropy can be achieved at the target confidence.

A stricter calibration accounts for the fact that entropy and top-1 confidence are coupled. 
Given a fixed top-1 confidence \(c\), the entropy is maximized when the remaining probability mass \(1-c\) is uniformly distributed over the other \(V-1\) tokens:
\begin{equation}
H_{\max}(c)
=
-c\log c
-
(1-c)\log\frac{1-c}{V-1}.
\label{eq:hmax_confidence}
\end{equation}
Substituting \(H_{\mathrm{ref}}=H_{\max}(c_{\mathrm{tar}})\) into Eq.~\eqref{eq:alpha_general} yields
\begin{equation}
\alpha_{\mathrm{strict}}
=
\frac{\tau-c_{\mathrm{tar}}}
{
-c_{\mathrm{tar}}\log c_{\mathrm{tar}}
-
(1-c_{\mathrm{tar}})\log\frac{1-c_{\mathrm{tar}}}{V-1}
}.
\label{eq:alpha_strict}
\end{equation}
Therefore, \(\alpha_{\mathrm{loose}}\) and \(\alpha_{\mathrm{strict}}\) define a theoretically motivated operating range:
\begin{equation}
\alpha \in 
\left[
\alpha_{\mathrm{loose}},
\alpha_{\mathrm{strict}}
\right].
\end{equation}
In practice, we choose \(\alpha\) from this interval rather than searching over a wide range. 
This ensures that the enhancement is strong enough to rescue moderately uncertain numerical or symbolic tokens, but not so strong that low-confidence tokens are released solely because of high entropy.

\mypara{Calibration of the repetition penalty weight \(\beta\)}
The role of \(\beta\) is different from \(\alpha\). 
Since \(r_m^t\) is shared by all masked positions in the same block, the repetition penalty does not change the relative ranking of candidate tokens. 
Instead, it uniformly lowers the release scores when the current block becomes too similar to the previous block. 
We calibrate \(\beta\) by specifying a target repetition ratio \(r_{\mathrm{tar}}\), above which borderline updates should be suppressed.

Let
\begin{equation}
s_{\mathrm{ref}}
=
c_{\mathrm{ref}}
+
\alpha \eta_{\mathrm{ref}} H_{\mathrm{ref}}
\end{equation}
be the score of a reference candidate before applying the repetition penalty. 
This reference candidate represents a token that would be released in a non-repetitive block but should be blocked once the block becomes overly repetitive. 
We impose the boundary condition
\begin{equation}
s_{\mathrm{ref}}-\beta r_{\mathrm{tar}}=\tau.
\label{eq:beta_boundary}
\end{equation}
Solving for \(\beta\) gives
\begin{equation}
\beta
=
\frac{s_{\mathrm{ref}}-\tau}{r_{\mathrm{tar}}}.
\label{eq:beta_general}
\end{equation}
Equivalently, if we denote the non-repetitive release margin by
\begin{equation}
\Delta_{\mathrm{ref}}
=
s_{\mathrm{ref}}-\tau,
\end{equation}
then
\begin{equation}
\beta
=
\frac{\Delta_{\mathrm{ref}}}{r_{\mathrm{tar}}}.
\label{eq:beta_margin}
\end{equation}

In our implementation, we set \(r_{\mathrm{tar}}=0.5\), meaning that when more than half of the current block repeats the previous block, borderline candidates should no longer be released by the adjusted score. 
This gives the simple calibration rule
\begin{equation}
\beta
=
2\Delta_{\mathrm{ref}},
\qquad
r_{\mathrm{tar}}=0.5.
\label{eq:beta_half_overlap}
\end{equation}
Thus, \(\beta\) is determined by the score margin we want to remove at the desired repetition level.

\section{Task-adaptive Mutation Budget on GSM8K}
\label{app:gsm8k}

\begin{table*}[t]
\centering
\caption{
Task-adaptive mutation-budget analysis on GSM8K using LLaDA2.0-Flash.
The standard decoding baseline achieves \(93.55\%\).
We vary the mutation bias \(\delta\), maximum mutated block index \(m_{\max}\), and mutation threshold \(\tau_{\mathrm{mut}}\).
Best results are bolded.
}
\label{tab:gsm8k_search_flash}
\setlength{\tabcolsep}{5pt}
\renewcommand{\arraystretch}{1.08}
\begin{tabular}{cc|cc|cc}
\toprule
\multirow{2}{*}{\(\boldsymbol{\delta}\)}
& \multirow{2}{*}{\(\boldsymbol{m_{\max}}\)}
& \multicolumn{2}{c|}{\(\boldsymbol{\tau_{\mathrm{mut}}=0.95}\)}
& \multicolumn{2}{c}{\(\boldsymbol{\tau_{\mathrm{mut}}=0.96}\)} \\
\cmidrule(lr){3-4}
\cmidrule(lr){5-6}
& & Acc. & \(\Delta\) & Acc. & \(\Delta\) \\
\midrule
\multicolumn{2}{c|}{Baseline}
& 93.55 & -- & 93.55 & -- \\
\midrule
\multirow{4}{*}{0.1}
& 1 & 93.56 & +0.01 & 93.71 & +0.16 \\
& 2 & 93.71 & +0.16 & 93.71 & +0.16 \\
& 3 & 93.56 & +0.01 & 93.56 & +0.01 \\
& 4 & \textbf{93.78} & \textbf{+0.23} & 93.71 & +0.16 \\
\midrule
\multirow{3}{*}{0.2}
& 1 & 93.71 & +0.16 & 93.71 & +0.16 \\
& 2 & \textbf{93.78} & \textbf{+0.23} & 93.78 & +0.16 \\
& 3 & 91.88 & -1.67 & 91.88 & -1.67 \\
\bottomrule
\end{tabular}
\end{table*}

We further analyze the mutation behavior on GSM8K in Table~\ref{tab:gsm8k_search_flash}. 
Different from AIME-style problems, GSM8K usually requires much shorter trajectories, with many examples completed within only \(4\)--\(5\) blocks. 
Therefore, the AIME-derived mutation budget \(m_{\max}=16\) becomes overly aggressive for GSM8K: mutation may affect almost the whole reasoning trajectory, leaving insufficient later blocks for the model to correct unstable perturbations. 
This explains why Selection remains helpful on GSM8K, while the original Mutation setting can become counterproductive.

After reducing the mutation budget to \(m_{\max}\in\{1,2,3,4\}\), Mutation becomes positive again. 
The best setting improves accuracy from \(93.55\%\) to \(93.78\%\), showing that mutation is still useful for GSM8K when restricted to early local exploration. 
More importantly, this result supports our interpretation that mutation should be matched to trajectory length: long competition-style problems benefit from a larger exploration window, while short arithmetic problems require a smaller and more conservative mutation budget. 
The absolute gain is small because GSM8K is already near saturation under the baseline, but the trend confirms that the negative effect does not come from mutation itself; rather, it comes from applying an AIME-scale mutation window to a much shorter reasoning task.

\section{Trajectory Analysis on AMIE-2025}
\label{app:tra_amie}

\subsection{Overall Trajectory Statistics}
\label{app:trajectory_table}

\begin{figure}
    \centering
    \includegraphics[width=0.94\linewidth]{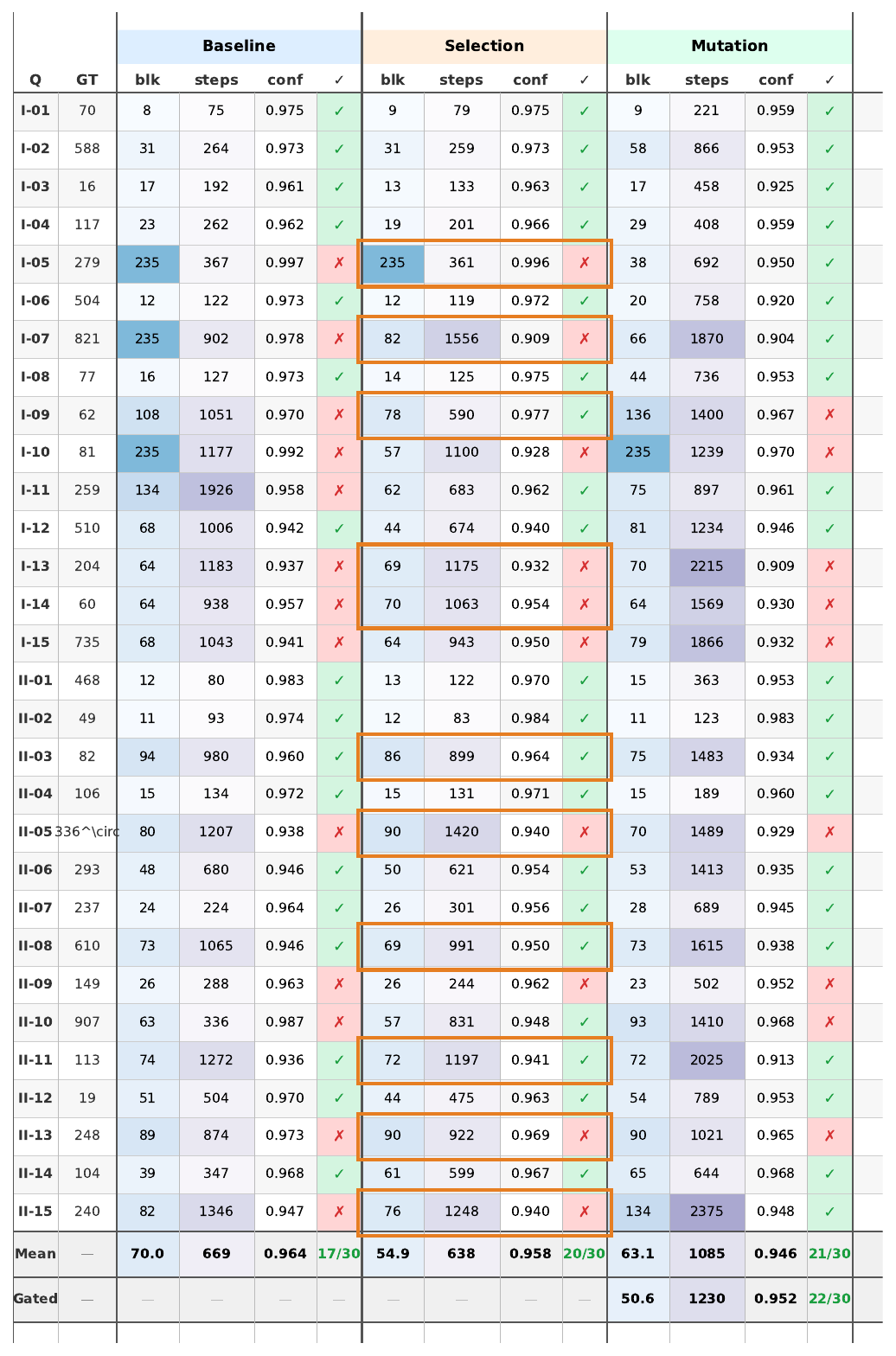}
    \caption{\textbf{Per-question trajectory statistics on AIME~2025.}
    For each problem, we compare Baseline, Selection, and Mutation in terms of the final block index, decoding steps, average confidence, and answer correctness. 
        The mean row reports the overall block/step cost and accuracy across all 30 AIME-I/II problems. 
        Selection reduces the average decoding steps while improving accuracy, suggesting that step-wise selection guides the trajectory toward correct answers more efficiently. 
        Mutation introduces additional diversity through four selectively mutated branches, which can improve correctness but may also perturb already correct trajectories. 
        The Gated setting applies mutation only after detecting likely selection failures, reducing unnecessary mutation cost and further improving accuracy.}
    \label{fig:table_aime}
\end{figure}

Figure~\ref{fig:table_aime} provides a question-level comparison of Baseline, Selection, and Mutation on AIME~2025. 
For each problem, we report the final block index, the total decoding steps, the average confidence, and whether the final answer is correct. 
This allows us to analyze not only the final accuracy, but also how different decoding strategies change the trajectory length and confidence behavior.

Overall, Selection improves both efficiency and accuracy. 
Compared with the baseline, the average block index is reduced from \(70.0\) to \(54.9\), and the average decoding steps are reduced from \(669\) to \(638\), while the number of solved problems increases from \(17/30\) to \(20/30\). 
This suggests that step-wise selection can guide the model toward correct answers with a more compact decoding trajectory, instead of allowing the generation process to drift into long and redundant continuations.

Mutation further increases the number of solved problems to \(21/30\), showing that structured perturbation can recover additional hard cases by introducing alternative numerical and symbolic continuations. 
However, unconditional mutation also increases the raw decoding cost, since multiple mutated branches are explored. 
This reveals a trade-off: mutation improves exploration diversity, but applying it to every problem may perturb trajectories that are already correct after selection.

To reduce this side effect, we further evaluate a gated mutation strategy. 
Instead of applying mutation to all trajectories, we first inspect the selection trajectory and trigger mutation only when the trajectory is likely to be wrong. 
Specifically, a trajectory is regarded as a likely failure if it reaches a late block index larger than \(64\) with low selection confidence below \(\tau_{\mathrm{mut}}\), or if the repetition ratio over the last \(10\) blocks exceeds \(0.5\). 
This strategy avoids unnecessary mutation on potentially correct trajectories. 
As shown in the Gated row, it further improves the solved count to \(22/30\), while reducing the average block index to \(50.6\) compared with unconditional mutation. 
This supports that mutation is more effective when used as a failure-aware corrective mechanism after selection.

\subsection{Block-wise Trajectory Visualization}
\label{app:trajectory_figures}

We further visualize six groups of block-wise trajectories on AIME~2025, covering AIME-I and AIME-II under Baseline, Selection, and Mutation. 
In each figure, the left axis denotes the average confidence of symbol/digit tokens, and the right axis denotes the number of decoding steps. 
Green curves indicate correct trajectories, while red curves indicate incorrect ones. 
These visualizations help explain how Selection and Mutation change the trajectories beyond the final accuracy numbers.

\begin{figure}
    \centering
    \includegraphics[width=1\linewidth]{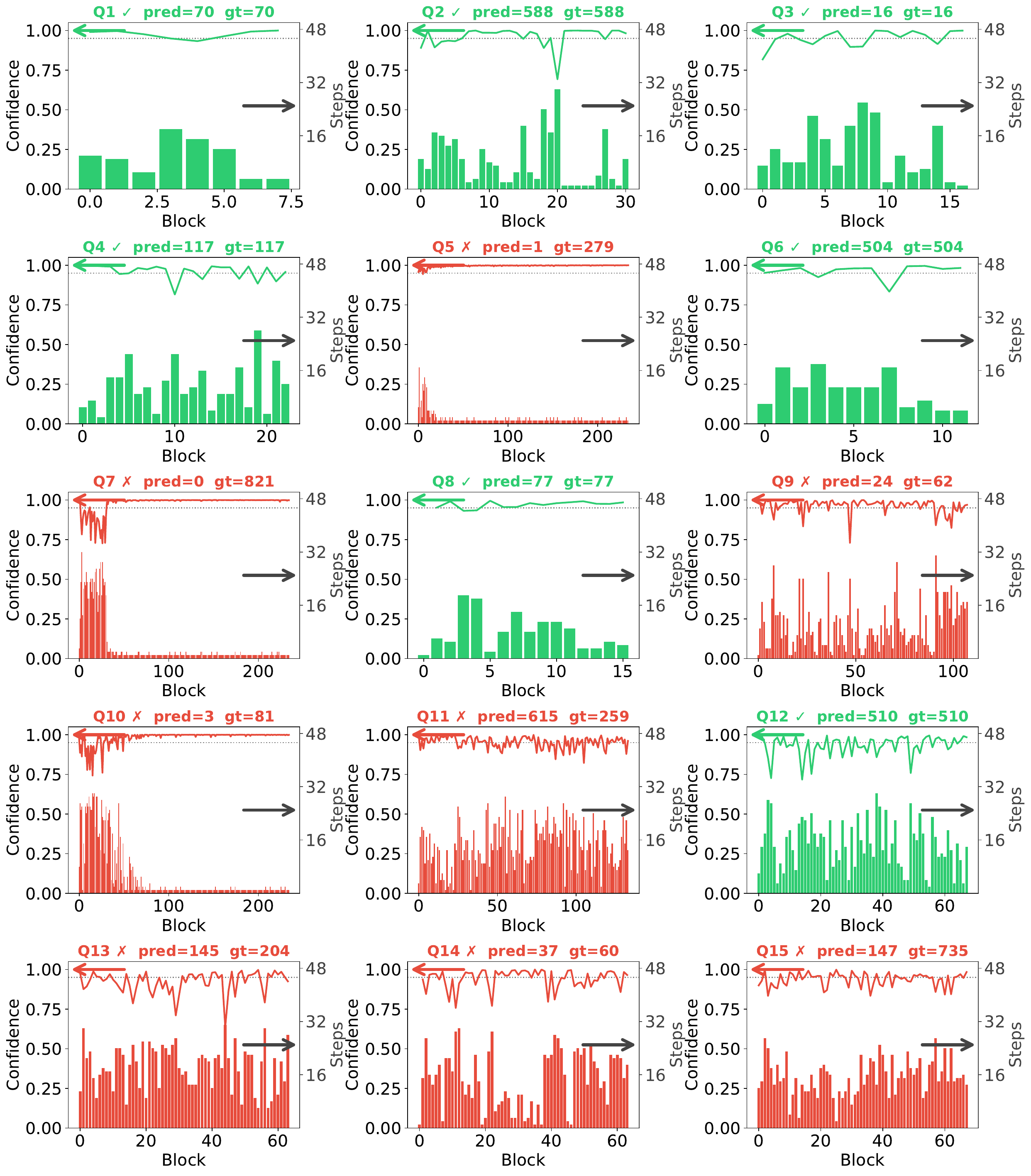}
    \caption{\textbf{Block-wise confidence and decoding-step analysis on AIME2025-I.}
    The left axis denotes the average confidence of symbol/digit tokens, and the right axis denotes the decoding steps. 
    Green curves indicate correct trajectories, while red curves indicate incorrect ones. 
    Successful trajectories are compact and stable, whereas failures mainly follow two modes: sampling-consistent failures stay highly confident but are accompanied by redundant natural-language filler, while sampling-sensitive failures exhibit larger block-wise fluctuations and lower symbol density, suggesting unstable transitions to low-information continuations.}
    \label{fig:b_aime1}
\end{figure}

\begin{figure}
    \centering
    \includegraphics[width=1\linewidth]{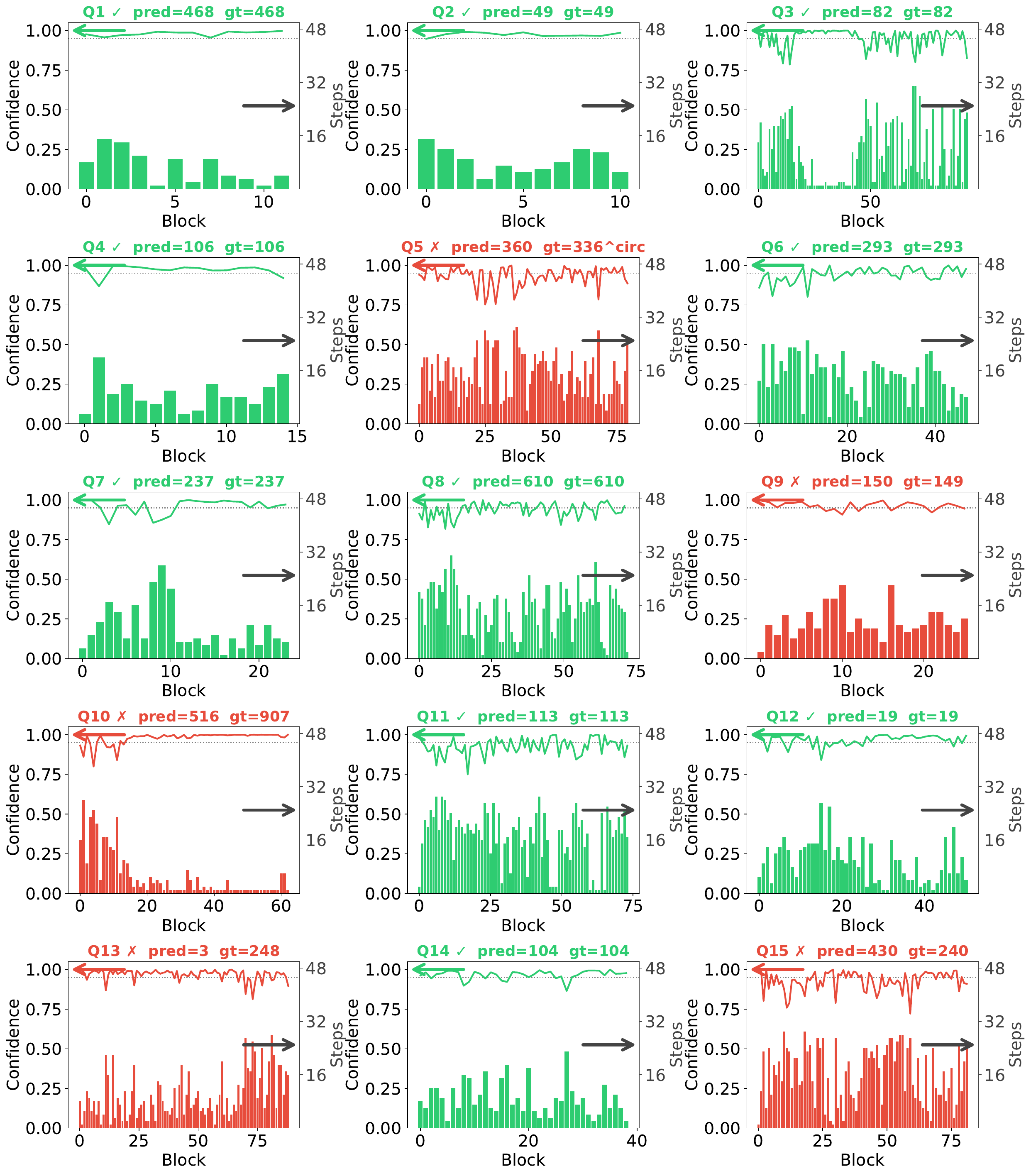}
    \caption{\textbf{Block-wise confidence and decoding-step analysis on AIME2025-II.}
    The left axis denotes the average confidence of symbol/digit tokens, and the right axis denotes the decoding steps. 
    Green curves indicate correct trajectories, while red curves indicate incorrect ones. 
    Successful trajectories are compact and stable, whereas failures mainly follow two modes: sampling-consistent failures stay highly confident but are accompanied by redundant natural-language filler, while sampling-sensitive failures exhibit larger block-wise fluctuations and lower symbol density, suggesting unstable transitions to low-information continuations.}
    \label{fig:b_aime2}
\end{figure}

\mypara{Baseline on AIME-I and AIME-II}
Figures~\ref{fig:b_aime1} and~\ref{fig:b_aime2} show the original confidence-based decoding trajectories. 
Successful trajectories are usually compact and stable: they reach valid answers within relatively short block chains and maintain smooth confidence evolution on symbol/digit tokens. 
In contrast, failed trajectories mainly exhibit two patterns. 
The first is a sampling-consistent failure mode, where the model remains highly confident on local symbol/digit predictions but produces redundant natural-language filler and fails to reach the correct answer. 
The second is a sampling-sensitive failure mode, where the block-wise confidence fluctuates more strongly and the symbol/digit density becomes lower, suggesting unstable transitions toward low-information continuations. 
These two modes motivate the need for both selection and mutation: selection stabilizes useful local decisions, while mutation provides alternative branches when the trajectory is trapped.

\begin{figure}
    \centering
    \includegraphics[width=1\linewidth]{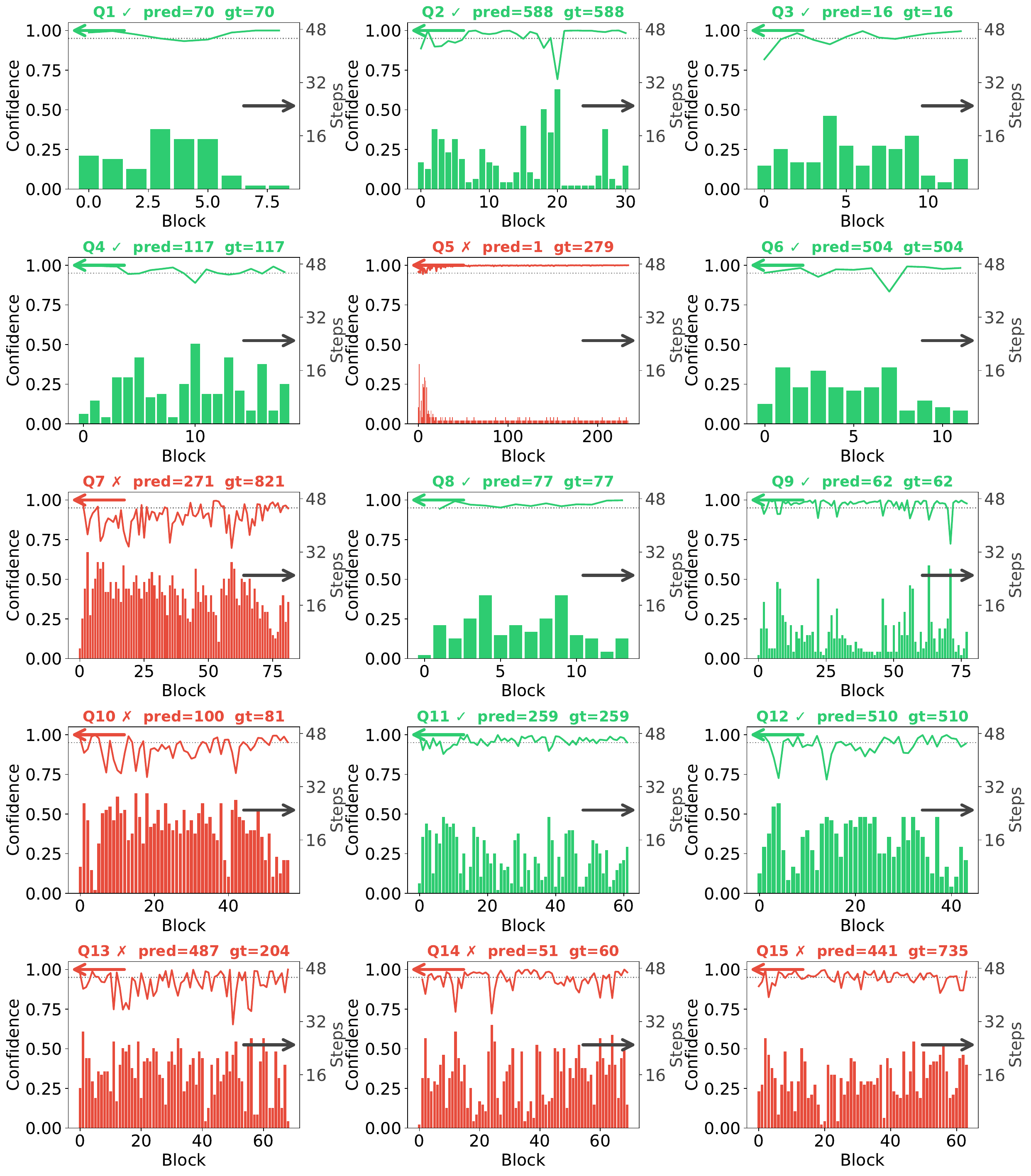}
    \caption{
    \textbf{Block-wise confidence and decoding-step analysis of Selection on AIME2025-I.}
    The left axis shows the average confidence of symbol/digit tokens, and the right axis shows the decoding steps.
    Green and red curves denote correct and incorrect trajectories, respectively.
    Compared with the baseline, Selection reduces the average block index from \(87.9\) to \(57.3\) and the average steps from \(709.0\) to \(604.1\), while recovering previously failed cases such as I-09 and I-11.
    This indicates that Selection leads to more compact and stable reasoning trajectories by suppressing redundant low-information continuations.
    }
    \label{fig:s_aime1}
\end{figure}

\begin{figure}
    \centering
    \includegraphics[width=1\linewidth]{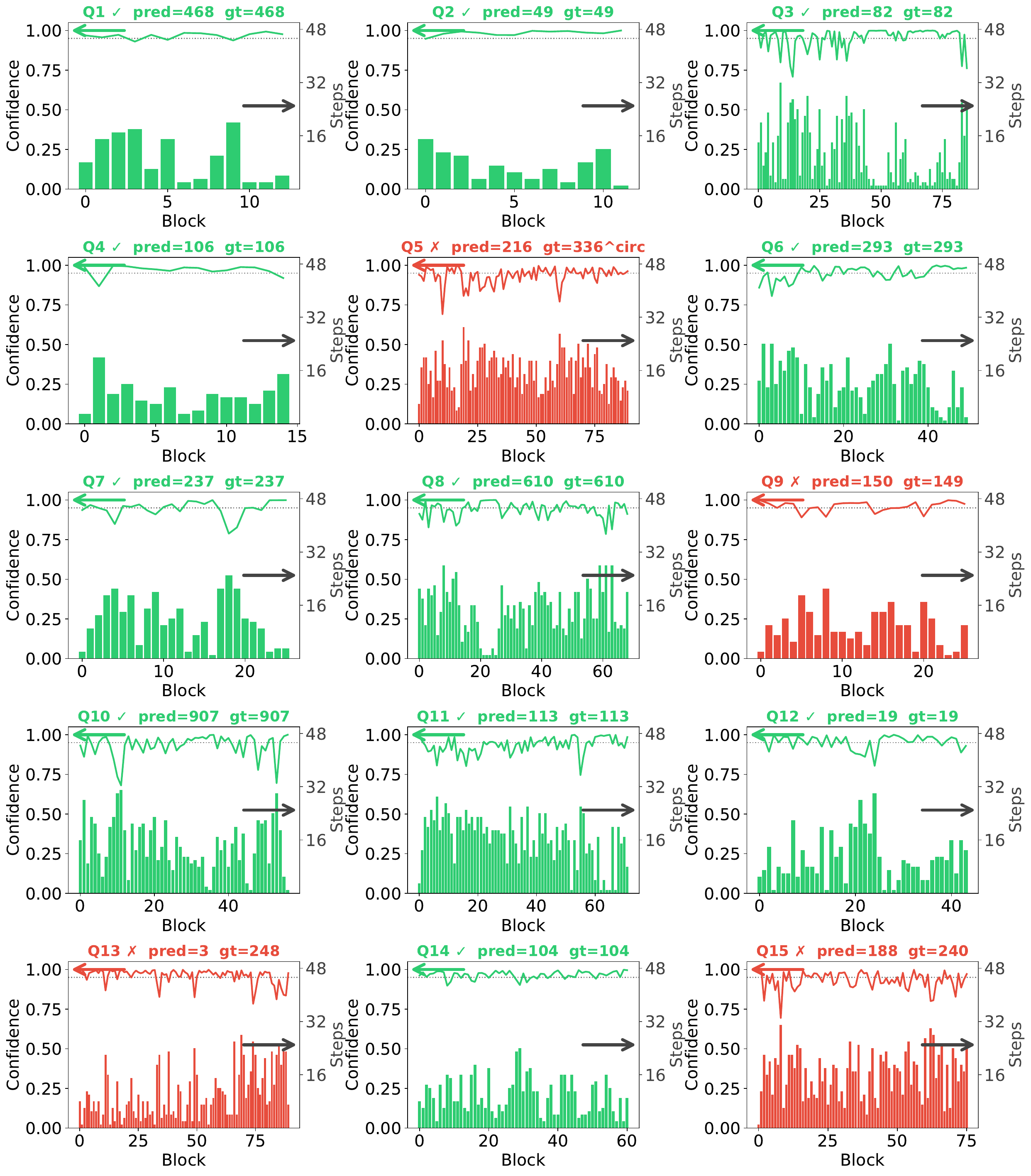}
    \caption{
    \textbf{Block-wise confidence and decoding-step analysis of Selection on AIME2025-II.}
    Compared with the baseline, Selection preserves similar average block confidence while improving the number of solved problems to \(11/15\).
    Several previously unstable trajectories are redirected to correct answers with more compact symbolic/numerical evolution, suggesting that step-wise selection helps suppress redundant low-information continuations and stabilizes late-stage reasoning.
    }
    \label{fig:s_aime2}
\end{figure}

\mypara{Selection on AIME-I and AIME-II}
Figures~\ref{fig:s_aime1} and~\ref{fig:s_aime2} show the effect of step-wise selection. 
Compared with the baseline, Selection produces shorter and more stable trajectories by prioritizing informative numerical and symbolic tokens. 
On AIME-I, several previously failed cases are redirected to correct answers, and the overall trajectory length is reduced. 
This indicates that Selection can suppress redundant low-information continuations and guide the model toward answer-bearing reasoning steps earlier. 
On AIME-II, Selection similarly improves the solved count while preserving stable confidence behavior, showing that the benefit is not limited to one subset of AIME~2025.

\begin{figure}
    \centering
    \includegraphics[width=1\linewidth]{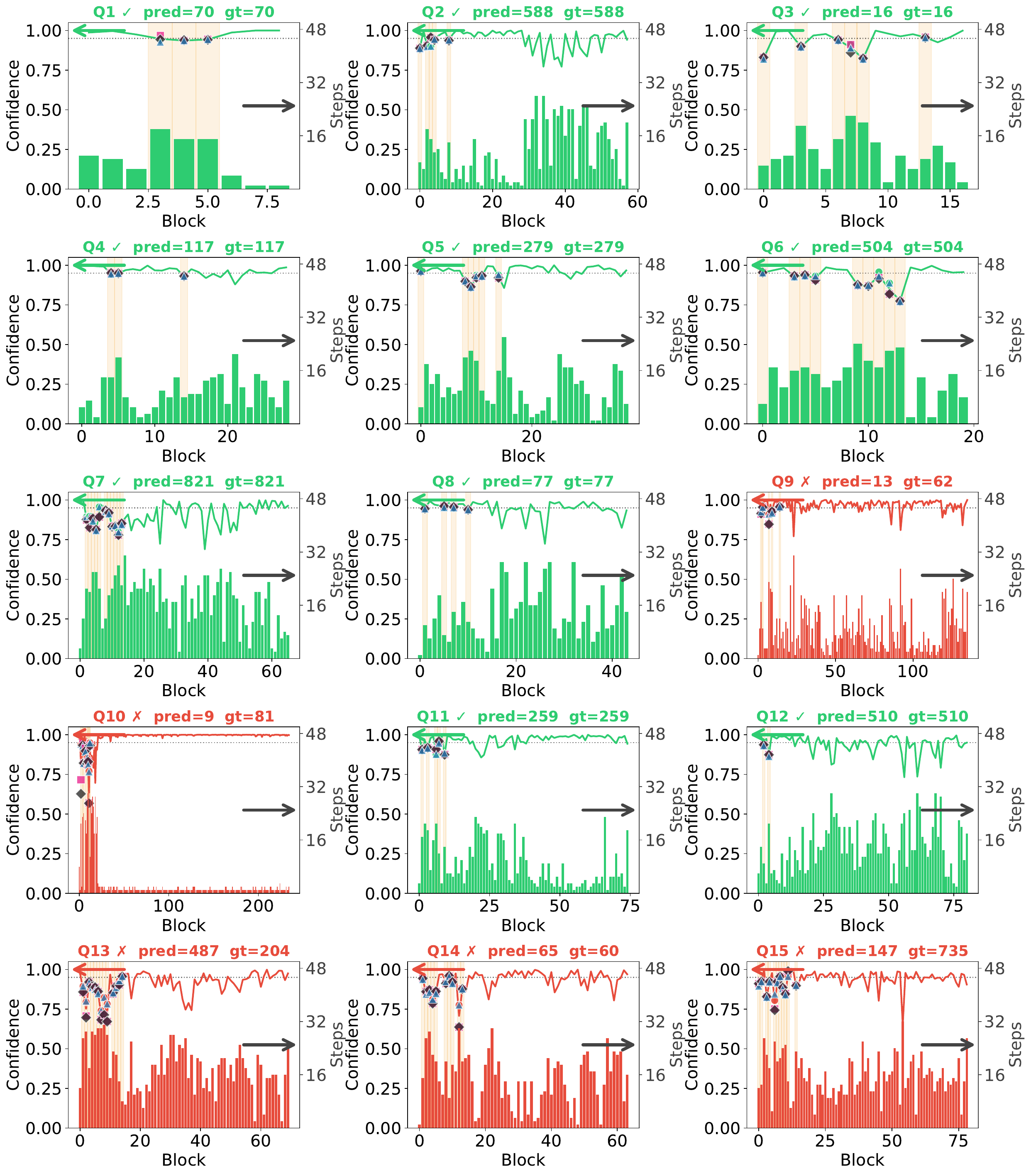}
    \caption{
    \textbf{Effect of Mutation over Selection on AIME2025-I.}
    Yellow-shaded regions indicate blocks within the first 16 block indices where structured mutation is applied.
    Compared with Selection, Mutation introduces more diverse trajectory evolution in the early stage, which helps escape some incorrect but stable reasoning paths and recover additional hard cases.
    This also shows that early selective mutation can increase exploration diversity, although it may introduce extra decoding cost.
    }
    \label{fig:m_aime1}
\end{figure}

\begin{figure}
    \centering
    \includegraphics[width=1\linewidth]{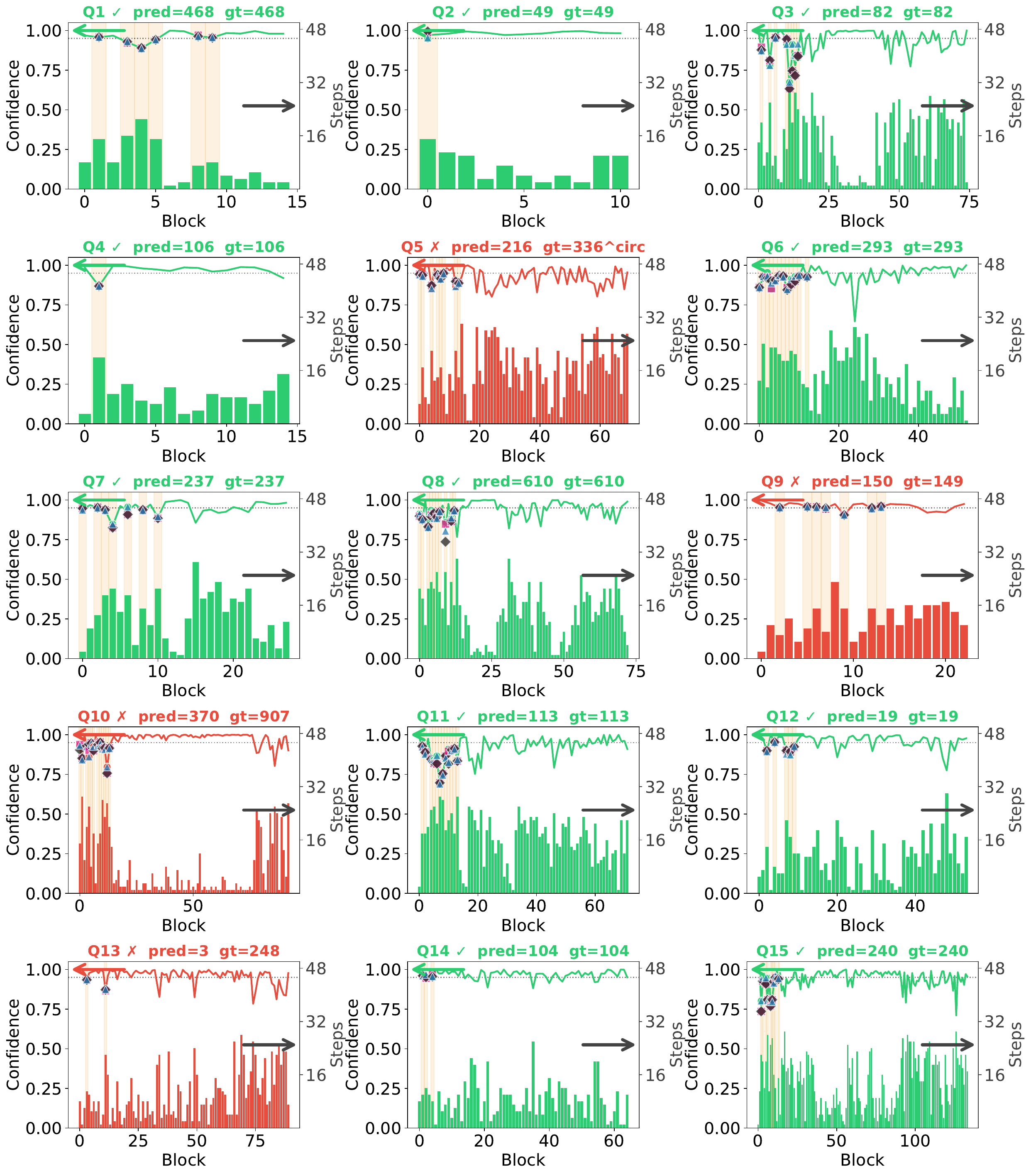}
    \caption{
    \textbf{Effect of Mutation over Selection on AIME2025-II.}
    Yellow-shaded regions denote the first 16 block indices where mutation is triggered.
    Compared with Selection, Mutation produces more diverse block-wise confidence trajectories, indicating that early structured perturbation effectively broadens the explored reasoning branches.
    While this added diversity does not always translate into immediate accuracy gain, it reveals the role of mutation in exploring alternative symbolic and numerical continuations.
    }
    \label{fig:m_aime2}
\end{figure}

\mypara{Mutation over Selection on AIME-I and AIME-II}
Figures~\ref{fig:m_aime1} and~\ref{fig:m_aime2} visualize the mutation trajectories built upon Selection. 
The yellow-shaded regions denote the early block indices where structured mutation is applied, limited to the first \(16\) block indices. 
Compared with Selection, Mutation introduces more diverse early-stage trajectory evolution, since numerical, symbolic, mixed, and neutral branches explore different local continuations. 
This diversity helps some trajectories escape incorrect but stable reasoning paths and recover additional correct answers. 
At the same time, the figures also show why mutation should not be applied blindly: for trajectories already corrected by Selection, mutation may introduce unnecessary exploration and increase decoding cost. 
This observation motivates the gated mutation strategy, where mutation is only triggered for trajectories that exhibit likely failure signals.

\section{Discussion and Future Work}
\label{app:discuss}

\mypara{Discussion}
Evolutionary Decoding improves mathematical reasoning through training-free test-time search over diffusion trajectories. 
Compared with standard decoding, it introduces additional inference cost because selection and mutation require extra trajectory-level operations. 
In our experiments, this cost is moderate, typically within a \(2\!\sim\!3\times\) increase in decoding time, while consistently improving pass@1 accuracy, pass@8 behavior, and trajectory stability. 
Thus, the extra computation provides a practical trade-off when higher reasoning reliability is needed.

Another design choice is using one unified hyperparameter configuration across all datasets and model variants. 
Although the parameters are selected and validated on AIME~2025 with LLaDA2.0-Flash, we directly transfer the same setting to other benchmarks and LLaDA2.0-mini without dataset- or model-specific tuning. 
This makes the evaluation conservative and reduces overfitting to individual benchmarks. 
For specific deployment scenarios or reasoning tasks, the guidance strategy could be further adapted to the task distribution for stronger performance.

Finally, the theoretical mechanism behind Evolutionary Decoding remains not fully understood. 
Our trajectory analyses show that selection improves compactness and stability, while mutation increases useful exploration diversity, but these observations do not yet fully explain why the strategy generalizes across datasets and model scales. 
This is partly due to the black-box nature of deep language models, where token confidence, trajectory evolution, and reasoning correctness are difficult to characterize analytically. 
The consistent success of ED suggests an interpretable mechanism in diffusion-style reasoning dynamics, and understanding when and why evolutionary operations improve diffusion decoding is an important future direction.

\mypara{Future Work}
Due to computation and evaluation constraints, we focus on mathematical reasoning benchmarks, where the final answers are well-defined and can be reliably checked. 
This setting allows us to clearly verify whether Evolutionary Decoding improves both final correctness and decoding trajectories. 
A natural future direction is to extend the analysis to broader reasoning tasks, such as code generation, scientific question answering, planning, and long-form problem solving. 
For these tasks, the informative token types may differ from mathematical reasoning: instead of numerical and symbolic tokens, useful guidance may involve code syntax, domain-specific entities, logical operators, or planning-related action tokens. 
Designing task-adaptive token guidance and more general mutation triggers is therefore an important next step.

We also plan to explore more efficient gated mutation strategies. 
Our current results show that mutation is most effective when applied after detecting likely selection failures, rather than being applied uniformly to all trajectories. 
A more general trigger could further reduce unnecessary computation while preserving the benefit of structured exploration. 
In addition, lightweight verifiers or process-level scoring functions may be incorporated into the survival stage to select candidate trajectories more accurately.

\mypara{Societal Impact}
This work may have positive impact by improving the reliability of mathematical reasoning in diffusion language models, benefiting education, scientific problem solving, and technical assistance. 
More reliable reasoning trajectories may provide clearer step-by-step explanations and support symbolic or numerical problem solving. 
Since the method is training-free, it can be applied to existing diffusion language models without additional training cost. 
Potential negative impacts include increased test-time computation and possible misuse of stronger reasoning capabilities. 
Moreover, improved decoding does not guarantee correctness in all cases, so appropriate verification remains necessary in high-stakes settings. 
The method does not involve private data, human subjects, or high-risk deployment, and our experiments are limited to public mathematical benchmarks.

\end{document}